%% file: acl_latex.tex
\pdfoutput=1
\documentclass[11pt]{article}

\usepackage[]{acl}

\usepackage{times}
\usepackage{latexsym}

\usepackage[T1]{fontenc}
\usepackage[utf8]{inputenc}

\usepackage{microtype}

\usepackage{inconsolata}

\usepackage{graphicx}
\usepackage{adjustbox}
\usepackage[capitalize]{cleveref}
\usepackage{booktabs}
\usepackage{multirow}
\usepackage{soul} % Used for highlighting

\usepackage{dcolumn}
\newcolumntype{d}{D{.}{.}{-1}}

\title{Attribute First, then Generate: \\ Locally-attributable Grounded Text Generation}

\author{Aviv Slobodkin$^{1}$\thanks{\hspace{5px}Equal contribution.} \quad Eran Hirsch$^{1\ast}$ \quad Arie Cattan$^{1}$ \quad Tal Schuster$^{2}$ \quad Ido Dagan$^{1}$ \\  $^{1}$Bar-Ilan University \qquad $^{2}$Google Research \\  
\texttt{\{lovodkin93, hirscheran, ariecattan\}@gmail.com} \\ \texttt{\qquad{talschuster@google.com} \qquad dagan@cs.biu.ac.il}
}

\definecolor{lightblue}{rgb}{0.0, 0.5, 0.9}

\usepackage{tcolorbox}
\tcbuselibrary{listings,breakable,skins}

\usepackage[notes=false, done=false, later=true, ]{dtrt}

\begin{document}
\maketitle
\input{sections/0_abstract}
\input{sections/1_introduction}

\input{sections/2_task}
\input{sections/3_modeling}

\input{sections/4_experimental_setup}
\input{sections/5_experiments}
\input{sections/7_related_work_arie}
\input{sections/8_conclusion}
\input{sections/9_limitations}
\input{sections/10_ethics}

% An ethics discussion is recommened but not mandatory

\section*{Acknowledgements}
This work was supported by the Israel Science Foundation (grant no. 2827/21).

\bibliography{anthology,custom}

\counterwithin{figure}{section}
\counterwithin{table}{section}

\input{sections/appendix}

\end{document}

%% file: sections/0_abstract.tex
\begin{abstract}
Recent efforts to address hallucinations in Large Language Models (LLMs) have focused on attributed text generation, which supplements generated texts with citations of supporting sources for post-generation fact-checking and corrections. Yet, these citations often point to entire documents or paragraphs, burdening users with extensive verification work.
In this paper, we introduce a \textit{locally}-attributable text generation approach, prioritizing concise attributions.
Our method, named \textit{``Attribute First, then Generate''}, breaks down the conventional end-to-end generation process into three intuitive steps: content selection, sentence planning, and sequential sentence generation.
By initially identifying relevant source segments (\textit{``select first''}) and then conditioning the generation process on them (\textit{``then generate''}), we ensure these segments also act as the output's fine-grained attributions (\textit{``select''} becomes \textit{``attribute''}). 
Tested on Multi-document Summarization and Long-form Question-answering, our method not only yields more concise citations than the baselines but also maintains—and in some cases enhances—both generation quality and attribution accuracy. Furthermore, it significantly reduces the time required for fact verification by human assessors.\looseness=-1\footnote{Our code is publicly available at \url{https://github.com/lovodkin93/attribute-first-then-generate}}

\end{abstract}

%% file: sections/1_introduction.tex
\begin{figure*}[t!]
\begin{adjustbox}{center}
    \centering
    \includegraphics[width=0.95\textwidth]{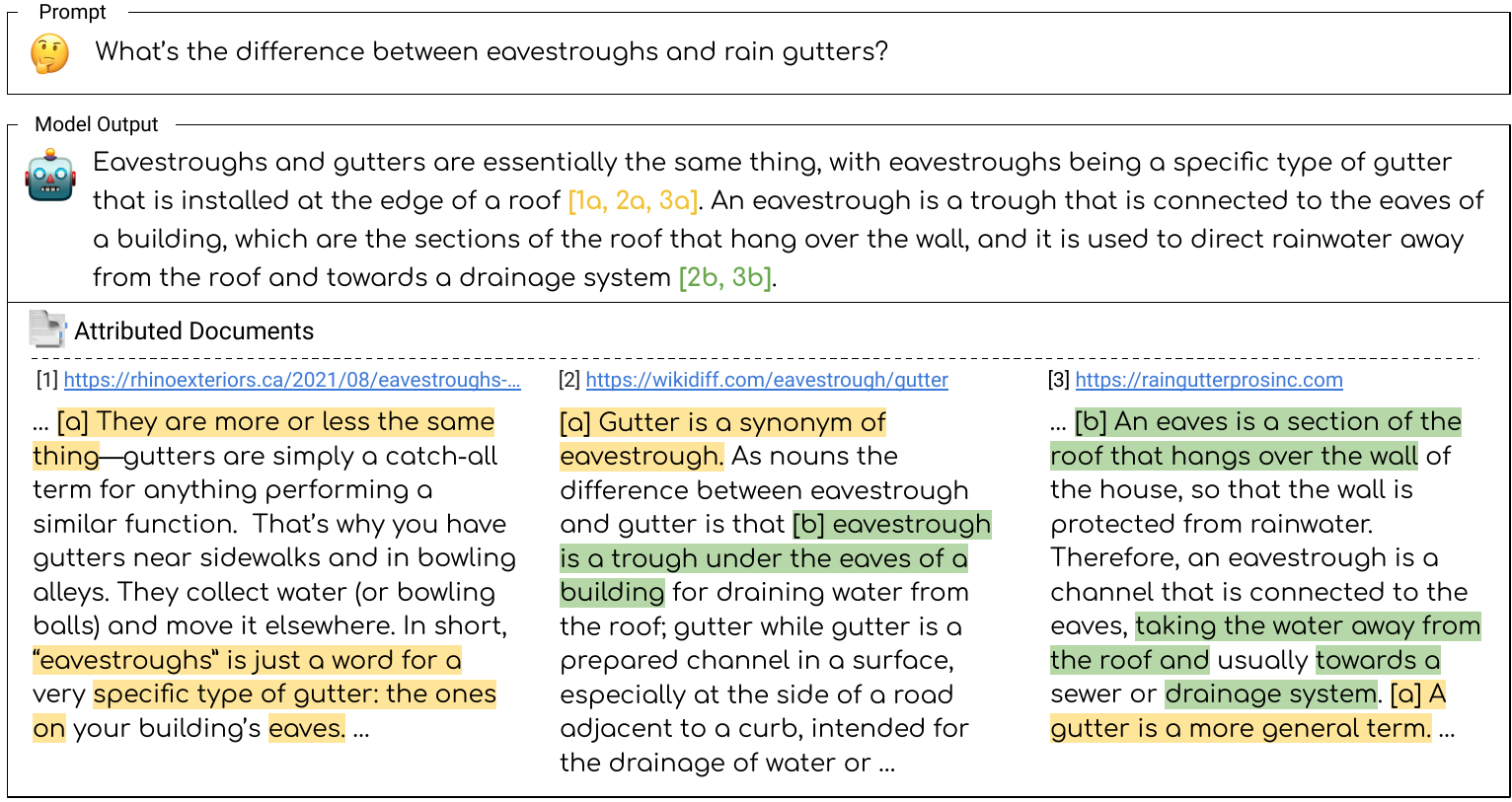}
\end{adjustbox}
    \caption{
    % Attributable generation for the QA text.
    % Given a question and three input documents, the model generates the answer accompanied with concise attributions only to relevant spans in the input texts that support the answer.
    In our approach, demonstrated here on LFQA, the input to the model assumes a question and a collection of documents, which could be retrieved or provided by the user. The generated text is grounded in these documents, accompanied with sentence-level citations to concise relevant spans from the supporting sources. Previous approaches \cite[e.g.][]{gao-etal-2023-rarr,gao-etal-2023-enabling} cite the whole source (truncated here for brevity), making it harder for readers to reliably verify the generation.
    }
    \label{fig:example}
\end{figure*}
% Source: https://docs.google.com/drawings/d/11rBalUhfAltvS-hptGOiVkPf51TsPr_ROoGjxXGzpWQ/edit

\section{Introduction}\label{sec:introduction}
Grounded text generation, which includes tasks like summarization and question-answering, aims to produce content that is derived from specific sources, either user-provided or retrieved via retrieval mechanisms \cite[e.g., RAG;][]{RAG}. To facilitate verification of models' adherence to these sources, recent years have seen a growing interest in \textit{attributed} text generation \citep{bohnet2023attributed}, which aims to create text alongside supporting evidence.
This method enhances models' credibility by enabling factuality verification. It also aids in detecting and addressing factual errors, a critical need given the observed frequency of such errors, termed ``hallucinations'', in model outputs compared to source texts~\citep{Mishra2024FinegrainedHD}.

While these attributions are aimed to facilitate factuality evaluation by focusing people's attention on relevant supporting texts, current approaches often yield rather \textit{coarse} attributions, pointing back to whole documents or paragraphs. 
Such attributions, though better than having none, require human assessors to exhaustively sift through many irrelevant details in the cited content, resulting in a time-consuming and somewhat ineffective fact-checking process. Can we do better?

% The main goal of these attributions is to make human assessment more tractable by restricting the focus only to the cited passages. Furthermore, attributions empower users to delve deeper into the facts presented in the model's outputs by exploring the cited sources. However, most existing works generate rather \textit{coarse} attributions, pointing back to entire documents or paragraphs. Such attributions, though more focused than the alternative, still require human assessors to exhaustively read many irrelevant details in the cited content, resulting in a time-consuming and somewhat ineffective fact-checking process. Can we do better?

In this work, we extend attributed text generation to also consider attribution conciseness, reformalizing the task as \textit{Locally}-attributable Text Generation (see \cref{sec:task}). 
We introduce a criterion for attributions to be \textit{precise}, targeting only the most relevant text snippets, from specific sentences down to sub-sentence spans, while avoiding non-essential details. 
This criterion complements the \emph{full coverage} requirement, which stipulates that every generated fact be backed by a cited snippet.
As illustrated in \cref{fig:example}, given a query and several documents, the model output includes citations to relevant spans within the source documents for each generated sentence instead of broadly attributing the entire documents~\citep{gao-etal-2023-enabling}. These spans represent the minimal set of relevant source snippets needed to cover the content of the output sentences. 

Following our localization criterion, we propose a novel attribution-driven generation approach, aimed to ensure both \textit{conciseness} and \textit{full coverage}, termed \textit{``Attribute First, then Generate''}. 
Rather than jointly producing text with citations~\citep{gao-etal-2023-enabling} or getting the attribution post-generation~\citep{bohnet2023attributed}, we propose to \textit{first} choose relevant details from the source text(s), and \textit{then} focus the entire generation process on them. Specifically, we decompose the conventional end-to-end grounded generation approach into three intuitive steps, as depicted in Figure~\ref{fig:architecture}: (1) content selection (choosing relevant details from source texts), (2) sentence-level planning (organizing and grouping content for sentence fusion), and (3) sentence-by-sentence generation (based on the selected and structured content). 
Notably, by dictating the generation process, the initially selected details de facto serve as attributions.
This ensures a closer attribution adherence to the generated output, as the entire generation process is now guided by these attributions. It also offers more precise supporting evidence, in the form of these selected details, thereby fulfilling our objectives of attribution accuracy and conciseness.

Our method, adaptable to various grounded generation tasks, requires only slight adaptations in the content selection step to cater to specific lexical needs, e.g., salience in summarization or query relevance in question-answering.
We showcase its effectiveness in both Multi-document Summarization (MDS) and Long-form Question-answering (LFQA),
exploring both finetuning and few-shot prompting strategies.
Evaluations across automatic and human assessments reveal that our strategy not only meets but in some cases surpasses existing baselines in generation quality while also achieving high attribution accuracy. Importantly, it also leads to significantly shorter citations, thereby reducing the effort needed for fact-checking by over 50\%.

Altogether, our work is the first to propose localized attributions and to develop a method geared to provide such concise citations. 
This level of granularity in attribution enhances the efficiency of the fact-verification process, by focusing users' attention on the most relevant details.
We thus suggest that future research will build upon our paradigm to further advance locally-attributed generation.

%% file: sections/2_task.tex
\begin{figure*}[t]
\begin{adjustbox}{center}
    \centering
    \includegraphics[width=0.8\textwidth]{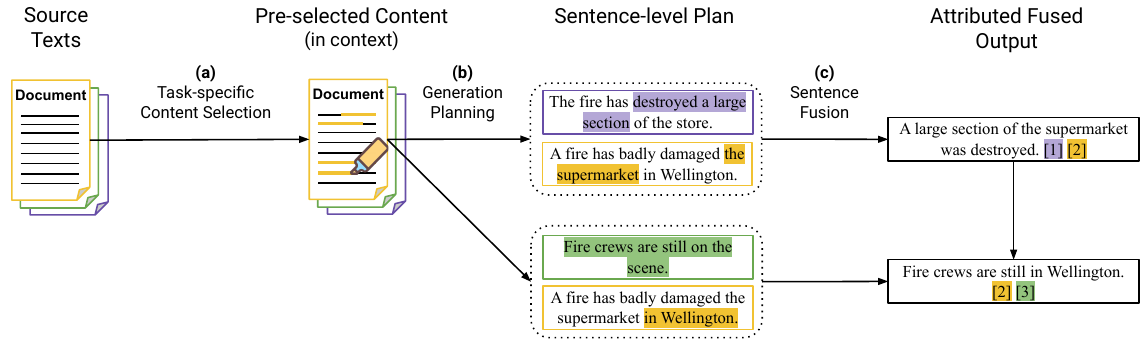}
\end{adjustbox}
    \caption{Our \emph{attribute first} process guides the model to output fluent texts that are consistent with input sources, and include fine-grained sentence-level attributions to localized text spans (i.e., highlights) from the inputs.}
    % \vspace{-0.2cm}
    \label{fig:architecture}
    
\end{figure*}
% Source: https://docs.google.com/drawings/d/1fF1NZZd4cIp5f9NZCZVxYG3-A4s4SfifsLAPayib0kc/edit

\section{Locally-attributable Grounded Text Generation}\label{sec:task}

The task of Locally-attributable Grounded Text Generation involves processing a given set of documents $\mathcal{D}$ to produce an output text $y$. The set of documents may either be provided by the output of a retrieval system or by the user. The output, consisting of sentences $s_1,\ldots,s_n$, should both fulfill the base underlying task (e.g., summarizing for MDS or generating a response for LFQA) and also include attributions $C_1,\ldots,C_n$ pointing back to supporting text from $\mathcal{D}$. For these attributions to effectively facilitate factuality verification by human readers, it is vital that they support as much of the output as possible, while also achieving the highest granularity in both the input and output.

Concretely, \textit{output} granularity 
entails assigning the smallest information units in the generated text $y$---in this work, individual sentences $s_i$---with a corresponding attribution set $C_i$, consisting of one or more supporting citations $c_i^j\in{C_i}$. 
Simultaneously, \textit{input} granularity requires each citation $c_i^j$ to point to a specific span of text $h_k\in\mathcal{D}$, possibly non-consecutive. Importantly, these spans should be as concise as possible while ensuring completeness in attribution, i.e., that every piece of information in the output, which in our setting refers to each output sentence $s_i$, is fully backed by the union of its citations $C_i$.
For example, in Fig.~\ref{fig:example}, $C_i$ is depicted as squared brackets at the end of each sentence with pointers to spans $h_k$, shown as colored highlights in $\mathcal{D}$ (we use ``highlights'' and ``spans'' interchangeably).

%% file: sections/3_modeling.tex
\section{Modeling}\label{sec:modeling}

This work aims to integrate source attribution with the content-grounded generation process, to achieve more targeted attribution. Our approach is based on uncovering the inherent decision-making mechanisms involved in text generation. 
We first describe our suggested \emph{Attribute First, then Generate} scheme in \cref{sec:scheme}. Then, we propose two strategies for this framework's application: a prompt-based in-context learning approach (\cref{sec:icl}) and fine-tuning designated components (\cref{sec:fine_tuned}). 
For each approach, we provide a brief overview of our implementation strategy for the framework's phases, while a more in-depth elaboration can be found in \cref{sec:implementation_details}.

\subsection{Attribute First, then Generate} \label{sec:scheme}

To reliably achieve the desirable fine-grained attribution, we introduce the \textit{Attribute First, then Generate} paradigm, outlined in \cref{fig:architecture}. This method is structured around three key steps, mirroring the inherent decision-making processes involved in text generation: content selection, sentence-planning, and sentence-by-sentence generation. 
Though applicable to any grounded text generation task, we specifically test its application in Multi-document Summarization (MDS) and Long-form Question-answering (LFQA),  adapting only the content selection step to each task's unique requirements.

\paragraph{Content Selection}
The approach begins with identifying relevant spans from the source that would contribute information to the generated output ((a) in \cref{fig:architecture}). 
These spans then function as a more focused grounding content, guiding the lexical choices in the generation process. Notably, as these chosen segments constitute the primary information for the resulting text, they effectively serve as its (fine-grained) attribution.

\paragraph{Sentence Planning}
This step is designed to achieve sentence-level attribution in the \textit{output} through the introduction of an intermediate sentence-level planning step ((b) in \cref{fig:architecture}). Here, the selected spans from the previous step are organized into coherently ordered clusters, where the spans grouped within each cluster naturally fit within the same sentence. This step, which breaks down the constrained generation process into smaller, more manageable steps, serves two purposes: 
it simplifies the model's task by conditioning each generation iteration on a single cluster rather than the entire set of selected spans, while also leading to the desired, sentence-level attribution in the \textit{output}, in the form of the corresponding cluster's highlights.

\paragraph{Sentence-by-sentence Generation}
In the final step ((c) in \cref{fig:architecture}), the model executes the text generation according to the outlined plan, adopting a sentence-by-sentence approach to preserve the built-in attribution from the previous steps. Specifically, it uses the next set of highlights $C_{i+1}$ and all preceding sentences $s_{1:i}$, to generate the next sentence $s_{i+1}$, by maximizing $p(s_{i+1} | s_{1:i}, C_{i+1})$. This ensures each sentence expresses the content of the designated highlights while also integrating seamlessly with the preceding output.

\subsection{In-context Learning} \label{sec:icl}
For each of the substeps delineated in the previous section, we propose a prompt-based in-context learning (ICL) strategy, where a model is guided with specific instructions and carefully selected few-shot examples. 
The content selection step ((a) in \cref{fig:architecture}) is formulated as $\mathcal{D} \rightarrow h_1,\ldots,h_m$, where $\mathcal{D}$ is the set of source documents and $h_i$ corresponds to a verbatim span from $\mathcal{D}$. Namely, a model is tasked with outputting relevant spans from $\mathcal{D}$, separated by a designated markup (see prompts in \cref{fig:MDS_content_selection_prompt,fig:LFQA_content_selection_prompt}).
In MDS, these spans are aimed to capture salient information, while for LFQA, the model, presented with a query, is instructed to identify spans answering that query.
These spans are then located within the source documents via string-matching, with any undetectable span being omitted (for more details, see \cref{subsec:implementation_details-content_selection}).
Notably, in all subsequent steps, we ``highlight'' the identified spans within the document set $\mathcal{D}$ via designated markups before and after each span, akin to the methodology used in prior work \citep{gehrmann-etal-2018-bottom, slobodkin-etal-2022-controlled, slobodkin2024multireview}.

For the sentence planning step ((b) in \cref{fig:architecture}), the documents, now ``highlighted'', are concatenated, truncating each at the final highlighted span to manage input length and focus the model's attention. 
To highlight, we use two designated markups, \textit{<highlight\_start>} and \textit{<highlight\_end>}, which we insert into the text before and after each consecutive `highlighted' span, respectively (see prompts in \cref{fig:MDS_clustering_prompt,fig:LFQA_clustering_prompt}).
The model is instructed to organize $h_1,\ldots,h_m$ into an ordered set of highlight clusters $C_1,\ldots,C_n$, where $C_i=\{h_k\}_{k=1}^{|C_i|}$, to guide the subsequent generation phase. For more details, see \cref{subsec:implementation_details-generation_planning}.

Finally, in step (c), the model is guided to generate the output one sentence at a time, adhering to the structured plan. Specifically, we separate the generation process into a series of model calls, each focusing on a single cluster's highlights, concatenated in the same manner as in step (b).
The model is also presented with all previously generated sentences $s_{1:i}$. The task is then structured as maximizing $p(s_{i+1} | s_{1:i}, C_{i+1})$, guiding the model to produce the subsequent sentence $s_{i+1}$, considering both the current cluster of highlights $C_{i+1}$ and the previously-generated sentences $s_{1:i}$.\footnote{In \cref{subsec:prefix_benefit} we show that conditioning on $s_{1:i}$ improves coherence.} 
For more details, see \cref{subsec:appendix_attributed_fused_output} as well as the full prompts structure in \cref{fig:MDS_sent_generation_prompt,fig:LFQA_sent_generation_prompt}.

\paragraph{Chain-of-Thought.} 
We also explore a variation that combines the sentence-planning and sentence-by-sentence generation stages ((b) and (c)), inspired by the Chain-of-Thought paradigm \citep[CoT;][]{wei2022chain}.
Given the selected content from step (a), the model is tasked with generating sentence-level plans, 
deciding on highlights to include in each synthesized sentence, alongside producing these sentences
(see prompts in \cref{fig:MDS_CoT_prompt,fig:LFQA_CoT_prompt}). 
The model is guided to produce the final output, after forming the plan, focusing on unifying and refining sentences for coherence.  For sentence-level attribution, we link each sentence in the output with the highlights from its planning stage in the CoT process.\footnote{We use spaCy sentencizer \cite{spacy2}.} For more details, refer to \cref{subsec:appendix_attributed_fused_output}.

\subsection{Fine-tuning} \label{sec:fine_tuned}
We also explore fine-tuning models for the various steps of our framework, aligning the input-output formulation of each subtask to its ICL counterpart in \cref{sec:icl}.
During content selection ((a) in \cref{fig:architecture}), our model is taught to copy task-specific segments from $\mathcal{D}$ verbatim, and to separate them using a special markup (see \cref{subsec:implementation_details-content_selection} and \cref{fig:ft_MDS_content_selection,fig:ft_lfqa_content_selection}). To further ensure strict lexical fidelity to the source text, a constrained decoding approach is adopted during inference (see \cref{appendix:logits_processor} for more details).

In the sentence planning step ((b) in \cref{fig:architecture}), we use special tokens to `highlight' the previously selected spans, and teach our model to copy verbatim the ``highlighted'' spans in a clustered and ordered manner, using special tokens to separate between different clusters and different highlights within a cluster (see \cref{subsec:implementation_details-generation_planning} and \cref{fig:ft_MDS_planning,fig:ft_lfqa_planning} for more details). 
This phase similarly incorporates a constrained decoding technique to maintain exact lexical matching to the highlighted content.
Finally, for the sentence generation step ((c) in \cref{fig:architecture}), we use special tokens to separate between the highlighted documents, the prefix, and for the LFQA setting the query (see \cref{subsec:appendix_attributed_fused_output} and \cref{fig:ft_MDS_sentence_generation,fig:ft_lfqa_sentence_generation}).

\paragraph{Joint Selection and Planning.} 
We also explore a joint strategy that combines the content selection and plan generation stages ((a) and (b)). Here, the entire document set $\mathcal{D}$ serves as input, and the output is an ordered series of ``highlight'' clusters $C_1,\ldots,C_n$. To ensure exact copying of source spans, we employ the aforementioned constrained decoding strategy. For more details, see \cref{subsec:implementation_details-generation_planning}.

%% file: sections/4_experimental_setup.tex
\section{Experimental Setup}\label{sec:experimental_setup}

\subsection{Datasets}\label{subsec:datasets}
To construct the training, development, and test splits for the different components within our framework, we employ source-output alignment datasets that map corresponding spans between inputs and outputs.
These alignments are adapted to suit the data requirements of each framework phase: for content selection, the alignments' \textit{source}-side spans serve as the target labels; in sentence planning, source spans corresponding to the same output sentence are clustered, with the order of output sentences determining the order of these clusters; for the sentence-by-sentence generation phase, the output sentences act as the target outputs for their respective source clusters.

For MDS, we use an alignment dataset derived from the DUC, TAC benchmarks\footnote{\url{https://duc.nist.gov/}} and the Multi-News dataset \citep{fabbri-etal-2019-multi}, comprising of pairs of summaries, corresponding source articles and summary-source alignments from \citet{ernst-etal-2022-proposition, ernst2024power}.
% .\footnote{\url{https://anonymous.4open.science/r/multi-news-dataset-5B4F} and \url{https://github.com/oriern/SuperPAL}.}
% \footnote{This dataset was annotated by some of this paper's authors for a different project, currently under peer-review. The link to the dataset is omitted for anonymity purposes.  For more details about the dataset, including statistics, see \cref{sec:dataset_preprocessing}.} 
For LFQA, we rely on human-annotated alignments of large language model responses with citations~\cite{liu-etal-2023-evaluating}. By filtering responses that were rated as verifiable, we ensure high-quality alignment for this setting. 
Further details and dataset processing information are provided in \cref{sec:dataset_preprocessing}.

% In our experiments, we use alignment datasets to derive designated train, dev, and test sets for the different subtask components.
% We explore two popular grounded generation settings - multi-document summarization (MDS) and long-form question-answering (LFQA). For MDS, we use the alignment dataset from \citep{ernst-etal-2021-summary} extracted from the Multi-News summarization dataset \citep{fabbri-etal-2019-multi} as well as the DUC and TAC summarization benchmarks.\footnote{\url{https://duc.nist.gov/}}
% For LFQA, we use the human verifiability evaluation from \citep{liu-etal-2023-evaluating}. In their work, the authors prompted several LLMs with a wide range of queries, producing responses with in-line citations after each sentence, followed by human evaluation of the citation's quality. Therefore, by filtering out those instances where the citations fully covered the corresponding output sentences, we managed to extract a high-quality alignment model for the LFQA setting.\footnote{For more details, see Appendix~\ref{sec:dataset_preprocessing}.}

\subsection{Models and Baselines}\label{subsec:models}
Our study investigates both ICL prompt-based LLMs and fine-tuned LMs.

\paragraph{In-context Learning.}
For our ICL approach (\cref{sec:icl}), we utilize the Gemini model \citep{team2023gemini}, applying designated few-shot prompts tailored to each subtask.\footnote{We used the Gemini API, which is free for academic purposes: \url{https://ai.google.dev/models/gemini}} 
For each setting, we also implement two baseline models for comparative analysis. For the first baseline, we prompt Gemini to perform the overarching task, namely summarization for MDS and answering a query for LFQA, end-to-end (see \cref{subsec:end-to-end-gen}). This baseline is primarily used to ensure the generated text's quality is not compromised in our framework. For the second baseline, we implement ALCE \citep{gao-etal-2023-enabling}, a prompt-based approach designed to enable the simultaneous generation of text and its attributions through in-line citations (e.g., [1] to cite source 1). See \cref{{subsec:ALCE_variant}} for more details.

\paragraph{Fine-tuned Models.}
For each of our proposed subtasks, we fine-tune models on our derived datasets (\cref{sec:fine_tuned}).  We choose to use \textsc{Primera} \citep{xiao-etal-2022-primera}, a Longformer Encoder-Decoder model \citep[LED;][]{https://doi.org/10.48550/arxiv.2004.05150} optimized for MDS, for both of our MDS and LFQA settings. 
Our choice of model stems from \textsc{Primera}'s leading performance in the MDS setting, attributed to its capability to handle long contexts and its pre-training technique.
Further, we found that \textsc{Primera} outperformed the Long-T5 large model \citep[Long-T5\textsubscript{\texttt{Large}};][]{guo-etal-2022-longt5} on our LFQA development set, where both models contain roughly the same number of parameters (Long-T5: 750M, \textsc{Primera}: 447M).
%leading to our decision to use it for this setting as well.
% Flan-T5 large model \citep[Flan-T5\textsubscript{large};][]{https://doi.org/10.48550/arxiv.2210.11416} for the LFQA setting.
Analogous to our ICL approach, we also fine-tune a variant to perform the corresponding task comprehensively for both settings.
% To acquire attribution for these baselines, we employ the RARR algorithm \citep{gao-etal-2023-rarr} to extract attributing snippets from input texts post-hoc, which we employ using Gemini  \citep{team2023gemini}.

\input{tables/tab_results}

\subsection{Evaluation}\label{subsec:evaluation}
Following our task definition in \cref{sec:task}, we aim to assess two aspects of the produced outputs. Firstly, we evaluate the generated texts' quality, focusing on coherence and task fulfillment—specifically, summary quality in MDS tasks and answer relevance in LFQA, consistent with standard evaluation practices in each respective task \citep{fabbri2021summeval, stelmakh-etal-2022-asqa}.
Secondly, we examine the attribution quality, looking at how well each sentence is supported by its accompanying citations and the conciseness of these citations (simply measured by the length of the cited texts). Our evaluation methodology combines automated metrics and human judgments.

\paragraph{Automatic Evaluation.}
To assess compliance with each task, we employ the \textsc{Rouge$_L$} \citep{lin-2004-rouge} and \textsc{BertScore} \citep{zhang2020bertscore} metrics. These compare generated texts with reference outputs, namely gold summaries for MDS and reference answers for LFQA.

For assessing citation conciseness, we report the average length (number of tokens) of the cited content per output sentence. 
To evaluate citation quality, we employ \textsc{AutoAIS}, an NLI-based metric for attribution evaluation, in line with prior studies \citep{bohnet2023attributed, gao-etal-2023-enabling} which have shown its correlation with human-judgment of attribution quality. 
Following those works, we employ the TRUE model from \citet{honovich-etal-2022-true-evaluating} for our NLI framework.\footnote{We also examined a more recent NLI model \citep[TrueTeacher;][]{gekhman-etal-2023-trueteacher}, and have found TRUE to correlate better with human judgment. See \cref{sec:correlation_with_human_judgement} for more details.}
The final \textsc{AutoAIS} score is the average number of sentences predicted as attributed by the classifier.
% To make the results comparable between approaches that cite full documents and those that cite text spans, we calculate two variants of \textsc{AutoAIS} which differ in their choice of source sentences used as premise. The first variation,  \textsc{AutoAIS$_D$}, considers the full source documents cited, discarding the information about specific sentences that were cited within the document. The second variation,  \textsc{AutoAIS$_H$}, considers only source sentences that have associated citations. The second variation is not calculated for approaches that cite entire documents.
% Next, for evaluating conciseness we report the sentence-level average citation length. 

% Despite its utility, \textsc{AutoAIS} tends to favor longer references (e.g., preferring citations to full documents), which may conflict with our goal of conciseness.
% Hence, we balance this metric with the average citation length and introduce a per-token citation value, calculated by dividing \textsc{AutoAIS} by citation length.
Lastly, we also report the average number of non-attributed sentences per instance.\footnote{Given the absence of reliable automatic metrics for fluency, we restrict its evaluation to the human evaluation, following standard practices \citep{fabbri-etal-2021-summeval}.}

\paragraph{Human Evaluation.}
The evaluation of generated attributed texts consists of two steps. First, annotators are asked to evaluate text coherence on a 5-point Likert scale, and in the LFQA setting also its helpfulness in addressing the query.
Next, annotators proceed to evaluate the quality of citations. We follow the sentence-level attribution evaluation protocol from \citet{gao-etal-2023-rarr}, which extends the \textit{Attributable to Identified Sources} \citep[AIS;][]{rashkin2022measuring} human evaluation framework of attribution to the sentence-level. Annotators judge whether each output sentence is fully supported by its attribution, scoring 1 for full support and 0 otherwise.
The final AIS score is the average of these scores across sentences. 
For more details, see \cref{sec:annotation_interface}.

%% file: tables/tab_results.tex
\begin{table*}
    \centering
    \small
    \adjustbox{max width=\textwidth}{
        \begin{tabular}{llcc|ccc}
            & Method & \multicolumn{1}{c}{\textsc{Rouge$_L$$\uparrow$}} & \multicolumn{1}{c}{\textsc{BertScore$\uparrow$}} & \multicolumn{1}{c}{\textsc{AutoAIS$\uparrow$}} & \multicolumn{1}{c}{\textsc{Length$\downarrow$}}& \multicolumn{1}{c}{\textsc{No Att. (\%)$\downarrow$}} \\
             \toprule
              \parbox[t]{2mm}{\multirow{4}{*}{\rotatebox[origin=c]{90}{ICL}}} & \textsc{Gemini} & 19.0 & \textbf{86.1}  & - & - & - \\
             & \textsc{ALCE} & \textbf{20.3}  & \textbf{86.1} & \textbf{88}.\textbf{7} & 843.6 & 3.4 \\
             % \textsc{RARR$_{\texttt{ICL}}$} & 19.0 & 86.1 & & 71.2 & 137.7 \\
             % \textsc{RARR$_{\texttt{fine-tuned generation}}$} & 19.9 & 85.8 & &  &  \\
             % \midrule
             & \textsc{Attr. First} & 16.7 & 85.2 & 79.5 & 92.7 & 0.0 \\
             % & \textsc{Attr. First}\textsubscript{\tאזexttt{recovery}} & \textbf{20}.\textbf{8} & \textbf{86}.\textbf{7} & 57.4 & 55.9 & 1.03 & 0.0 \\
             & \textsc{Attr. First}\textsubscript{\texttt{CoT}} & 19.5 & \textbf{86.1} &  72.8 & 75.3 & 1.5 \\
             \midrule
             % \emph{Finetuned models} \\
             \parbox[t]{2mm}{\multirow{3}{*}{\rotatebox[origin=c]{90}{FT}}} & \textsc{Primera} & 19.9 & 85.8 & - & - & - \\
             & \textsc{Attr. First} & 16.5 & 84.9 & 64.4 & \textbf{33.1} & 0.0 \\
             & \textsc{Attr. First}\textsubscript{\texttt{joint}} & 18.0 & 85.3 & 48.9 & 57.1 & 0.0 \\

             \bottomrule
        \end{tabular}
    }
    \vspace{-4pt}
    \caption{MDS results of both in-context learning (ICL) and fine-tuned (FT) models. The left two columns are metrics that measure the quality of the output, ignoring attributions. The right columns report the attribution quality, including the length (number of tokens) of the cited text. \textsc{No Att.} shows the percentage of generated sentences lacking any attribution.}
    \label{tab:automatic_results_mds}
\end{table*}

\begin{table*}
    \centering
    \small
    \adjustbox{max width=\textwidth}{
        \begin{tabular}{llcc|ccc}
             & Method & \multicolumn{1}{c}{\textsc{Rouge$_L$$\uparrow$}} & \multicolumn{1}{c}{\textsc{BertScore$\uparrow$}} & \multicolumn{1}{c}{\textsc{AutoAIS$\uparrow$}} & \multicolumn{1}{c}{\textsc{Length$\downarrow$}} & \multicolumn{1}{c}{\textsc{No Att. (\%)$\downarrow$}} \\
             \toprule
             \parbox[t]{2mm}{\multirow{4}{*}{\rotatebox[origin=c]{90}{ICL}}} & \textsc{Gemini} & 33.1 & 89.6  & - & - & - \\
             
             & \textsc{ALCE} & 35.2 & 89.9 & 49.8 & 2153.3 & 26.9 \\
             % \textsc{RARR$_{\texttt{ICL}}$} & 33.1 & 89.6 & & 52.5 & 149.0 \\
             % \textsc{RARR$_{\texttt{fine-tuned generation}}$} & 32.2 & 88.9 & & &  \\
             
             & \textsc{Attr. First} & 35.8 & 90.5 & 78.7 & 65.2 & 0.0 \\
             % & \textsc{First Attr.$_{recovery}$} & 37.4 & 90.5 & 41.8 & 49.4 & 0.84 & 0.0 \\
             % & \textsc{Attr. First}\textsubscript{\texttt{recovery}} & \textbf{38}.\textbf{9} & \textbf{91}.\textbf{8}  & 72.0 & 49.4 & 1.45 & 0.0 \\
             & \textsc{Attr. First}\textsubscript{\texttt{CoT}} & \textbf{38.6} & \textbf{90.7} & \textbf{89}.\textbf{3} & 48.2 & 0.0 \\
             \midrule
            \parbox[t]{2mm}{\multirow{3}{*}{\rotatebox[origin=c]{90}{FT}}} &  \textsc{Primera} & 32.2 & 88.8 & - & - & - \\
            & \textsc{Attr. First} & 24.7 & 88.0 & 52.7 & \textbf{20.9} & 0.0 \\
             & \textsc{Attr. First}\textsubscript{\texttt{joint}} & 27.0 & 88.1 & 44.9 & 33.1 & 0.0 \\
             \bottomrule
        \end{tabular}
    }
    % \vspace{-4pt}
    % \caption{Results for the long-form QA dataset derived from \citet{liu-etal-2023-evaluating} annotations.}
    \caption{LFQA results of both in-context learning (ICL) and fine-tuned (FT) models. The left two columns are metrics that measure the quality of the output, ignoring attributions. The right columns report the attribution quality, including the length (number of tokens) of the cited text. \textsc{No Att.} shows the percentage of generated sentences lacking any attribution.}
    % \vspace{-0.2cm}
    \label{tab:automatic_results_qa}
\end{table*}

%% file: sections/5_experiments.tex
\section{Results and Analyses}\label{sec:results}

We provide results of our automated evaluation in \cref{tab:automatic_results_mds,tab:automatic_results_qa} and human evaluation in \cref{tab:human_results_mds,tab:human_results_qa}. 
Considering the resource-intensive aspects of human evaluation, we limit this assessment to the ALCE baseline and our paradigm's most efficient model, the ICL variant \textsc{Attr.\ First}\textsubscript{\texttt{CoT}}, chosen for its optimal balance between task performance and attribution accuracy, as indicated by the automatic evaluation.

We first note that our approach, which extracts focused spans from the input documents, leads to significantly shorter attributions than the ALCE baseline, where citations coarsely point to entire documents. For example, the \textsc{Attr.\ First}\textsubscript{\texttt{CoT}} variant's citations are on average 45 times shorter than ALCE's.\footnote{See examples of fine-grained attributions generated by our approach in \cref{sec:fine-grained_attribution_examples}.}
Importantly, we find that assessing our paradigm's attribution takes annotators roughly half the time compared to those of ALCE, emphasizing the benefit of attribution localization in streamlining the manual fact-checking process.

In terms of attribution accuracy, our method surpasses the ALCE baseline in the LFQA domain (both on the automatic \textsc{AutoAIS} and the human \textsc{AIS} metrics), while in MDS, although ALCE shows higher accuracy, it leads to twice as many \textit{un}attributed sentences.
We also note that while in the MDS setting our method may decrease attribution accuracy compared to ALCE's full-document citations, readers can always fall back to examining the full document if the provided highlights are insufficiently supporting.
Indeed, analyses of the human annotations revealed that around 42\% of the 109 sentences deemed unsupported were partially supported, with the remainder of the attributing details frequently found in immediate proximity to the highlighted sections (usually within the same or adjacent sentences).\footnote{See partially-attributed examples in \cref{sec:partially-attributing-examples}.} This observation indicates that while our sentence fusion step occasionally includes slightly more information than the selected highlights (later used as fine attributions), it mostly adds details from the surrounding context.\footnote{This trend is also observed in our automatic evaluation, where extending sub-sentence attributing spans to their corresponding full sentences improves \textsc{AutoAIS}. See \cref{sec:full_sentence_autoais}.} 
Hence, while suggesting room for improvement in strict adherence to highlights, this does not significantly hinder fact-checking effectiveness.\looseness=-1

\input{tables/tab_human_evaluation}

Importantly, the promising results on attribution do not compromise the quality of outputs. 
Our automatic analyses indicate minimal impact on summary quality in MDS and even an improvement in LFQA, as confirmed by the \textsc{Rouge$_L$} and \textsc{BertScore} metrics.
These results are further evident in our human evaluation, both in terms of the output's fluency and (for LFQA) in helpfulness for answering the query.
Notably, our in-context variants tend to produce higher-quality outputs than fine-tuned models.
This apparent advantage, observed also in attribution accuracy, is likely due to Gemini's stronger ability to generalize to long-context settings, as well as its larger model size.

Overall, our experiments show that our approach leads to orders of magnitude shorter attributions, with little impact on the attribution accuracy and the output's quality. 
We leave further improvements in fine-tuning the individual designated components to future research.

\begin{figure}[t]
\begin{adjustbox}{center}
    \centering
    \includegraphics[width=0.8\columnwidth]{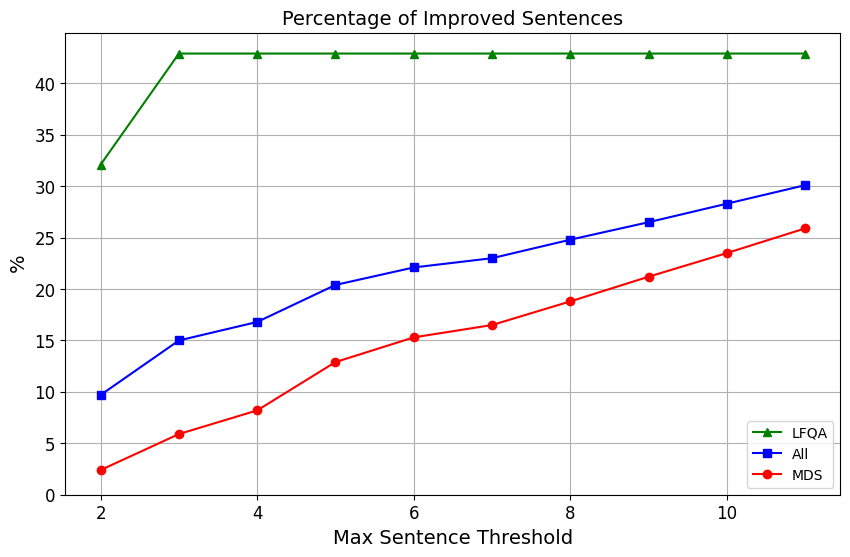}
\end{adjustbox}
    \caption{Percentage of more cohesive sentences when including the prefix in the input across varying upper limits of sentence index in the output.}
    \label{fig:prefix_vs_no_prefix_thr}
    
\end{figure}

\subsection{Do Models Benefit from Seeing the Prefix?}\label{subsec:prefix_benefit}
We also examine whether providing models with the context of previously generated sentences indeed aids the generation process, particularly in enhancing output cohesion as defined by \citet{maimon2023novel}. Cohesion involves sentences being \textit{``referentially linked''}, e.g., via co-referring mentions, or linked through discourse connectors. 
To analyze the impact of the prefix on cohesion, we modified the sentence generation prompt to exclude the prefix and conducted the sentence fusion step (step (c) in \cref{fig:architecture}) using the ICL variant for both MDS and LFQA tasks, maintaining identical sentence-level plans. 
Starting from the second sentence to isolate the impact of the prefix, we examined 20 instances per task. Our analysis of 113 sentences reveals that approximately 30\% were more cohesive when including the prefix. Specifically, 23\% showed improved referential linkage, mostly attributed to better use of co-reference, and 7\% demonstrated better discourse connector usage.\footnote{See \cref{sec:prefix_vs_no_prefix} for examples.} 
Additionally, our findings suggest that prefixes play an increasing role in cohesion as the output progresses (see \cref{fig:prefix_vs_no_prefix_thr}), implying that cohesion gains importance as the text develops.\footnote{The observed plateau in cohesion levels in LFQA is attributed to its outputs generally not exceeding three sentences.}

%% file: tables/tab_human_evaluation.tex
\begin{table}
    \centering
    \small
    \adjustbox{max width=\columnwidth}{
        \begin{tabular}{lccc}
             Method & \multicolumn{1}{c}{Fluency$\uparrow$} & \multicolumn{1}{c}{AIS$\uparrow$} & \multicolumn{1}{c}{Time (sec)$\downarrow$} \\
             \toprule
             \textsc{ALCE} & 4.7 & \textbf{89.3} & 47 \\
             \textsc{Attr. First}\textsubscript{\texttt{ICL-CoT}} & \textbf{4.9} & 79.9 & \textbf{22} \\
             \bottomrule
        \end{tabular}
    }
    % \vspace{-4pt}
    \caption{Human evaluation on the MDS outputs. Fluency evaluates the text coherence. AIS assesses the text's attribution support for each sentence, with full support scoring 1, and 0 otherwise. Time is a measurement of the amount of seconds it took an annotator to verify if a text is fully supported by its attribution.}
    \label{tab:human_results_mds}
    \vspace{-4pt}
\end{table}

\begin{table}
    \centering
    \small
    \adjustbox{max width=\columnwidth}{
        \begin{tabular}{lcccc}
             Method & \multicolumn{1}{c}{Fluency$\uparrow$} & \multicolumn{1}{c}{Helpful.$\uparrow$} & \multicolumn{1}{c}{AIS$\uparrow$} & \multicolumn{1}{c}{Time (sec)$\downarrow$} \\
             \toprule
             \textsc{ALCE} & \textbf{4.9} & \textbf{4.7} & 87.6 & 59 \\
             \textsc{Attr. First}\textsubscript{\texttt{ICL-CoT}} & \textbf{4.9} & 4.5 & \textbf{94.4} & \textbf{35} \\
             \bottomrule
        \end{tabular}
    }
    % \vspace{-4pt}
    % \caption{Results for the human evaluation for the QA dataset derived from \citep{liu-etal-2023-evaluating} annotations.}
    \caption{Human evaluation on the LFQA outputs. Metrics are similar to \cref{tab:human_results_mds}. The helpfulness metric measures the helpfulness of responses to queries on a 5-point Likert scale.}
    \vspace{-4pt}
    \label{tab:human_results_qa}
\end{table}

%% file: sections/7_related_work_arie.tex
\section{Related Work}
\label{sec:related_work}

A few recent works have proposed several methods to generate text accompanied by attribution, with varying levels of granularity in the source and the generated text. In Long-Form Question Answering (LFQA), the LaMDA system \citep{thoppilan2022lamda} provides attribution for the \textit{entire} response in the form of URLs pointing to \textit{entire} documents. In GopherCite \citep{menick2022teaching}, the \textit{entire} response is attributed to fine-grained snippets from the documents. ALCE \citep{gao-etal-2023-enabling} provides attribution for each generated sentence to one or multiple \textit{entire} input documents. 
SEMQA~\citep{schuster2023semqa} is a recent multi-source semi-extractive dataset for LFQA, which encourages models to integrate factual spans copied verbatim from the input, along with free-text connectors to ensure coherence. This method is by design fine-grained both at the source and the response level using string matching. Yet, this semi-extractive paradigm is limited in scenarios requiring a higher level of abstraction, such as Multi-document Summarization (MDS).
In contrast to all those approaches, our localized attribution achieves granularity both in the input and the output, without restricting the model to extractive text generation.
 
In information seeking, recent works have proposed to find attribution in post-hoc, by retrieving documents and matching each part of the answer to specific pieces of text from the retrieved documents~\citep{bohnet2023attributed, gao-etal-2023-rarr}. These methods face a combinatorial challenge because the relevant snippets of text in the source documents need to be determined.

The field of attributed text generation is part of a broader endeavor to address hallucinations in models' outputs and increase their overall trustworthiness. In this context, much work has been done on characterizing the sources of hallucinations \citep{dziri2022origin, rawte2023exploring,yao2023llm, li2024dawn} as well as on strategies to identify and mitigate them \citep{azaria-mitchell-2023-internal, manakul2023selfcheckgpt, min-etal-2023-factscore, slobodkin-etal-2023-curious, mishra2024finegrained}, thereby advancing the reliability of model-generated content.

Another related line of research focuses on text-planning for improving output quality and reducing hallucinations~\citep{gehrmann-etal-2018-bottom, narayan-etal-2021-planning, narayan-etal-2023-conditional,zhao-etal-2020-bridging}, by introducing an additional step of planning before the final text generation. 
This area of study includes efforts to decompose the generation process into distinct phases, with some works focusing on the content-selection step \citep{gehrmann-etal-2018-bottom, cho-etal-2019-improving, ernst-etal-2021-summary}, while others focus on the content consolidation phase \citep{lebanoff-etal-2019-analyzing, lebanoff-etal-2020-learning, slobodkin-etal-2022-controlled, slobodkin-etal-2023-dont, hirsch-etal-2023-revisiting}. 
Our work also leverages this planning paradigm to improve the factuality of the output. However, unlike previous methods, our pipeline is carefully designed to also provide high-quality fine-grained attributions.
Perhaps most related to our method is the work by \citet{ernst-etal-2022-proposition} that divides
% cite Ruben's paper 
the MDS process into detecting salient propositions, clustering them, and transforming them into distinct sentences. However, their method results in outputs that resemble bullet-point summaries, consisting of single-proposition sentences, rather than producing more fluid, multi-proposition sentences typical of natural text. In contrast, our method maintains the coherence and natural flow of generated texts without being restricted to single-proposition sentences.

Text-planning is also related to the Chain-of-Thought (CoT) approach that demonstrated improved performance on math problems~\citep{wei2022chain}, coding and programming puzzles~\citep{chen2021evaluating,schuster2021p3}, and other reasoning tasks~\citep{saparov2023language}. Both approaches build on the idea of decomposing the generation task to allow an initial reasoning and planning stage, before generating the predicted output. In CoT, the planning and output prediction are typically performed in a single generation call, whereas in text-planning the planning stage is separated and could be performed either by the same LM or using different models and operations. Therefore, CoT is more suitable for in-context learning whereas text-planning works well with finetuned specialized models. We explore both options in this work.

% Grounded text generation is an area focusing on generating text from source documents. Among its tasks, one can find summarization~\citep{nallapati2016abstractive, nallapati2016classify, paulus2017deep, gehrmann2018bottom}, where the input consists of one or more documents, and the output is expected to cover the salient information within those documents. Another well-researched task is Long-Form Question-Answering \citep{fan2019eli5, stelmakh_asqa_2023}, 

% includes tasks like long-form question-answering \citep{fan2019eli5, stelmakh_asqa_2023}, summarization \citep{nallapati2016abstractive, nallapati2016classify, shapira2020massive, brazinskas-etal-2020-unsupervised, zhao2022mrs}, and dialogue systems \citep{yan2017building, xu2019end, thoppilan_lamda_2022}, with most related datasets aimed at end-to-end training \citep{fan2019eli5, bravzinskas2020few, liu2021durecdial, iso2022comparative}.

%% file: sections/8_conclusion.tex
\section{Conclusion}\label{sec:conclusion}

In this work, we present \textit{``Attribute First, then Generate''}, a novel scheme for attributable grounded text generation, comprised of separating the generation process into multiple steps, such that attribution is achieved as a by-product of the generation. To the best of our knowledge, we are the first to provide such fine-grained attributions in both the generated text and the references. Through automatic and manual evaluations, we showed that our decomposed generation pipeline either matches or outperforms existing baselines 
in terms of task-specific metrics. Importantly, our approach yields orders of magnitude shorter attributions, shortening the fact-checking time by almost 50\%, with little impact on the attribution accuracy.
While this work focused on a particular implementation of content selection and generation, we encourage future explorations to extend our \textit{``Attribute First, then Generate''} paradigm to other configurations.
We hope that our research will motivate future work in this direction, improving the quality of text generation while also providing helpful concise attributions to external sources, thereby increasing models' trustworthiness and safe downstream use.

%% file: sections/9_limitations.tex
\section*{Limitations}\label{sec:limitations}
Our Attribute First, Then Generate approach tends to use more computing resources and is also slower compared to direct end-to-end generation, as detailed in \cref{sec:latency_and_tkn_count}. This is because it requires several steps, each involving a separate call to the model.

Furthermore, our pipeline is comprised of several interlinked components, which can potentially lead to error propagation. However, this potential issue is partially alleviated by the inherent robustness of the individual components. For instance, should the content selection mechanism identify a broad array of text fragments, including some that are irrelevant, subsequent stages in the process are designed to filter out such non-essential information.

%% file: sections/10_ethics.tex
\section*{Ethics Statement}\label{sec:ethics_statement}

\paragraph{Crowdsourcing.} For our human evaluation we used the Amazon Mechanical Turk\footnote{\url{https://worker.mturk.com/}} (MTurk) crowdsourcing platform. 
Recruitment of annotators was conducted via email invitations dispatched from the MTurk platform, aimed at individuals who had previously demonstrated high performance in similar research endeavors. 
Compensation was provided not only for the actual annotation tasks but also for an initial onboarding phase, which included familiarization with the annotation guidelines and going over feedback from the authors. 
The pricing for each annotation instance was calibrated based on an estimated completion time, aiming to maintain a compensation rate of approximately 12\$ per hour. 
Additionally, we closely monitored the actual time required for task completion and offered post-hoc reimbursements in compensation for instances where the required time exceeded our initial estimate by more than 5\%.

\paragraph{Datasets licenses.} The DUC and TAC datasets were acquired according to the required NIST guidelines (\url{duc.nist.gov}), and we do not publish any data from these datasets. The annotation data from \citet{liu-etal-2023-evaluating} is published with the MIT license. The Multi-News dataset \citep{fabbri-etal-2019-multi} is published with a non-commercial license intended for research and educational purposes. Accordingly, our human evaluation annotations will also be limited to non-commercial use. 

\paragraph{AI assistants}

AI assistants were exclusively employed to enhance the grammatical structuring of the text.

%% file: sections/appendix.tex
\appendix
\section{Datasets Preprocessing}\label{sec:dataset_preprocessing}

Datasets statistics are reported in  \cref{tab:datasets_sizes}.
For LFQA, we utilize published human annotations of LLM responses with in-line citations \citep{liu-etal-2023-evaluating}. We use the existing train, dev, and test splits, and we apply a few filters on this dataset as follows. Responses from LLMs can contain statements that are not supported by the retrieved documents and are annotated with ``Partial Support'' or ``No Support''. We classify each statement as supported by the input documents if either one of its citations is annotated with ``Full Support'' or if the union of all citations is annotated with ``Full Support''. We remove examples from the dataset containing unsupported statements. In addition, the dataset also includes human annotations for perceived utility (on a Likert scale between 1 to 5). We only keep examples with perceived utility 4 and 5 (``Strongly Agree'' and ``Agree''). Lastly, the dataset contains evidence excerpts for each generated statement, which we use as citations. We remove examples from the dataset if we cannot align the evidence excerpt with the source documents.

For the training processes in the MDS setting, we use the alignment dataset from \citet{ernst-etal-2022-proposition}, automatically extracted from the DUC and TAC summarization datasets.\footnote{\url{https://duc.nist.gov/}} For the development and test splits, we use the alignment dataset from \citet{ernst2024power}, extracted from the Multi-News summarization dataset \citep{fabbri-etal-2019-multi}. The dataset consists of 100 summaries with a corresponding set of source news articles, comprising an average of 3 articles per summary. Additionally, for each such pair of summaries and set of documents, the authors trained qualified annotators to identify aligning spans, following the approach delineated in \citet{slobodkin-etal-2022-controlled}.

\input{tables/tab_datasets_sizes}

% \section{Details about the Alignment Dataset in the MDS Setting}\label{sec:mds_alignment_dataset_details}

% \input{tables/oracle_full}

% \section{Oracle Experiment Results}
% \label{appendix:oracle_results}

% Results for our oracle experiment are provided in \cref{tab:oracle_results_mds,tab:oracle_results_lfqa} for the MDS and LFQA settings, respectively.

% \input{tables/hyperparam_tuning_backtracking}

% \input{tables/hyperparam_tuning_MDS_e2e}
% \input{tables/hyperparam_tuning_MDS_alce}
% \input{tables/hyperparam_tuning_MDS_content_selection}
% \input{tables/hyperparam_tuning_MDS_clustering}
% \input{tables/hyperparam_tuning_MDS_FiC_CoT}
% \input{tables/hyperparam_tuning_MDS_backtracking}
% \input{tables/hyperparam_tuning_MDS_iterative}

% \input{tables/hyperparam_tuning_LFQA_clustering}
% \input{tables/hyperparam_tuning_LFQA_content_selection}

% \input{tables/hyperparam_tuning_LFQA_alce}

\input{tables/correlation_with_human_judgement}

\section{Meta-Evaluation for \textsc{AutoAIS}}\label{sec:correlation_with_human_judgement}
Prior work that suggested using NLI models for automatic attribution evaluation \citep{bohnet2023attributed, gao-etal-2023-enabling, gao-etal-2023-rarr} has specifically advocated using the TRUE model \citep{honovich-etal-2022-true-evaluating}. Subsequent developments have introduced an enhanced NLI model, named TrueTeacher \citep{gekhman-etal-2023-trueteacher}.
In our study, to determine the most appropriate NLI model for the automatic assessment of attribution within the \textsc{AutoAIS} framework, we conduct a correlation analysis between the outcomes of these models and our human-generated annotations.
We also evaluate an alternative method that averages the models' probabilities for the entailing label ('1') across all sentences, rather than binary outcomes.
To calculate correlation, we use both Kendall-Tau and Spearman's rank correlations, as suggested in \citep{deutsch2022re}.
We also apply bootstrapping \citep{efron1987better} by performing 1000 samplings of 150 instances (with repetition) and calculating correlation scores for each such subset. We report the average correlation and 95\% confidence intervals for each metric.
Our findings, presented in \cref{tab:correlation_with_human_judgement}, indicate that despite the advancements offered by the TrueTeacher model, the TRUE model still aligns more closely with human judgment.
Furthermore, the approach of averaging probabilities for the label '1' correlates significantly worse for both models. Consequently, we opt to continue utilizing the TRUE model for our \textsc{AutoAIS} evaluations.

\input{resources/MDS_content_selection_prompt}

\input{resources/MDS_clustering_prompt}

\input{resources/MDS_sent_generation_prompt}

\input{resources/MDS_CoT_prompt}

\input{resources/LFQA_content_selection_prompt}

\input{resources/LFQA_clustering_prompt}

\input{resources/LFQA_sent_generation_prompt}

\input{resources/LFQA_CoT_prompt}

\input{resources/FT_MDS_examples_io}

\input{resources/FT_LFQA_examples_io}

\section{Implementation Details}\label{sec:implementation_details}

In this section, we describe the implementation details for all the component described in \cref{sec:modeling}, both the ICL and fine-tuning approaches. \textsc{Rouge$_L$} and \textsc{BertScore} were calculated with the evaluate library \footnote{\url{https://huggingface.co/docs/evaluate/en/index}} \citep{lin-2004-rouge}.

For our ICL experiments, we use the Gemini-Pro model \citep{team2023gemini}, last updated on December 2023, which is available via API calls.\footnote{\url{https://ai.google.dev/models/gemini}}
In all LFQA experiments, the query is included as part of the input for each component. For all subtasks and baselines, we perform a grid search on the optimal number of in-context examples (1-3 in the MDS setting and 1-4 in the LFQA setting) and temperature hyperparameter (0.1,0.3,0.5,0.7,0.9) by evaluating each combination on the respective development set.
For all fine-tuned variants, we fine-tune \textsc{Primera} for both tasks. We also tested training \textsc{LongT5}, but found \textsc{Primera} performs significantly better. The number of parameters in the \textsc{Primera} model is 447M. All fine-tuned models were trained for 10 epochs and the best model checkpoint was chosen by \textsc{Rouge$_L$}, unless mentioned otherwise. The training of each model took approximately 1 hour on one Nvidia A100 80GB GPU. Hyperparameter training included 3 epochs for each model and used the following sweep configuration: learning rate (max: 5e-4; min: 5e-8), warmup steps (max: 300; min: 0), weight decay (0.0, 0.2, 0.5).
\subsection{Content Selection}\label{subsec:implementation_details-content_selection}
% eran - write here the implementation details of the ft models

\paragraph{ICL}

In the ICL method, we present the model with the source texts and guide it to copy specific segments from those texts (see \cref{fig:MDS_content_selection_prompt,fig:LFQA_content_selection_prompt} for prompt examples in the MDS and LFQA settings, respectively). We use a designated \textit{``<SPAN\_DELIM>''} markup as a delimiter between such consecutive spans, for better parsing.
Within the MDS framework, the model is tasked with identifying salient excerpts, whereas in the context of LFQA, the model is also provided with a query,
 and is asked to find the parts of the texts that are relevant to answering this question. 
 Further, following our analyses of each setting's development sets, we restrict the model to an average and maximum number of words to choose (an average of 200 and 100 words and a maximum of 900 and 200 words for MDS and LFQA, respectively).
 Lastly, to manage source texts that exceed the model's input size limit, we truncate the fewest necessary sentences from those texts to ensure they fit within the specified limit.\footnote{We use spaCy \cite{spacy2} to separate documents into sentences.} 
After performing hyperparameter tuning, in the MDS we set $T=0.1$ and the number of in-context examples to 2, while in LFQA we set $T=0.3$ and the number of examples to 4.

\paragraph{Fine-tuned}
Examples are provided in \cref{fig:ft_MDS_content_selection,fig:ft_lfqa_content_selection}.
The input to the model is the set of all documents, concatenated with a special markup ``<doc-sep>'', which was also used in pre-training in \textsc{Primera}. To fit the input in a context length of 4096, we trim each document to the length of 4096 divided by the number of documents in each example. The output is a set of highlights separated with a new special markup ``<highlights-separator>''. We run hyperparameter-tuning on each setting based on the development set. The hyperparameters used for MDS are warmup steps of 229, weight decay of 0.5, and a learning rate of 4e-4. Conversely, the hyperparameters used for LFQA are warmup steps of 234, weight decay of 0, and a learning rate of 1e-4. We chose the best model checkpoint by calculating Intersection-over-Union (IoU) between the gold highlights spans and the predicted highlights spans.

\subsection{Generation Planning}\label{subsec:implementation_details-generation_planning}

\paragraph{ICL}
In the ICL setting, we present the model with the source text where we insert the mark-ups \textit{``<highlight\_start>''}  and \textit{``<highlight\_end>''} before and after each pre-selected span, (see \cref{fig:MDS_clustering_prompt,fig:LFQA_clustering_prompt} for the MDS and LFQA settings, respectively). We also enumerate the highlights across all documents, and for the LFQA setting, we also add the query. 
The model is directed to cluster highlighted segments that are likely to co-occur in a sentence in the same order it would then merge them into an output.
We discovered that providing the model with task-specific motivation enhances its performance; for instance, in MDS, the objective is to fuse highlights into a summary, whereas in LFQA, it aims to formulate a response to the query. 
To facilitate the parsing of the model's response, we encourage it to generate its response as a list of single-value dictionaries, with ``cluster'' as the key and a list of indices corresponding to the highlights within the current cluster as the value.
For hyperparameter tuning, we divide the gold ``highlights'' into pairs.
Pairs that correspond to the same target output sentence are categorized as positive, whereas others are considered negative. Subsequently, we evaluate the F1 score to assess the alignment between the predicted and gold clusters. This involves calculating the True Positives, False Negatives, and False Positives for the highlights pairs, based on the predicted clusters. 
Following our analyses, we opt for 2 examples and $T=0.5$ in the MDS setting, and 1 example and $T=0.1$ in the LFQA setting.

% For the plan recovery approach, we also take two steps - fusing all highlights and then pairing post-hoc between the output's sentences and the highlights. For the second step, we use the same model as in the finetuned models. Alternatively, for the first step, we show the model the ``highlighted'' source texts, akin to our aforementioned clustering prompt, and ask the model to generate a coherent output covering all and only the highlighted spans (see \cref{fig:MDS_regular_FiC_prompt}). As before, we find that giving motivation to the overall task improves the model's performance. We use 2 in-context examples and $T=0.3$ for the MDS setting, while for LFQA we find that 3 examples and $T=0.3$ work best on the development set.

\paragraph{Fine-tuned}
Examples are provided in \cref{fig:ft_MDS_planning,fig:ft_lfqa_planning}.
There are two variants for generation planning: generative clustering and joint. In the generative clustering approach, the input to the model is the set of all highlights, concatenated with a special markup ``<highlight-separator>'', which is the same as the output from the content selection model. The output is then a set of clusters, where each cluster is separated by a special markup ``<cluster-separator>'', and inside each cluster we have a set of highlights in the same format as the input. We run hyperparameter-tuning on each setting based on the development set. The hyperparameters used for MDS are warmup steps 0, weight decay 0, and learning rate 5e-5. The hyperparameters used for LFQA are warmup steps 298, weight decay 0, and learning rate 2e-4. At inference time, we used a beam size of 1. We use the same logits processor described in \cref{subsec:implementation_details-content_selection} to enforce the model to generate valid highlights.

In the joint approach, the input to the model is the original source documents, similar to the input in the content selection model. In contrast, the output is the clustered highlights. Similar to the content selection model, we use the same logits processor to enforce generating valid highlights. The hyperparameters used for LFQA are warmup steps 298, weight decay 0, and learning rate 2e-4.

\subsection{Attributed Fused Output}\label{subsec:appendix_attributed_fused_output}

\paragraph{ICL}
For the ICL variant, we iteratively show the model only the current cluster's highlights, in the same manner, we do so in the generation planning ICL experiments (see \cref{subsec:implementation_details-generation_planning}). Additionally, we present the model with what it generated to that point and guide it to generate the next sentence in the paragraph, which should cover all and only the highlighted spans (see \cref{fig:MDS_sent_generation_prompt,fig:LFQA_sent_generation_prompt}). We use 3 examples and $T=0.3$ for the MDS setting, and 2 examples with $T=0.1$ for the LFQA setting.

For the Chain-of-Thought approach, which merges generation planning with the fusion steps, 
we present the model with the ``highlighted'' source texts and ask it to fuse those highlights into a coherent output. Additionally, we guide the model, both in the instructions and through in-context demonstrations to iteratively cluster highlights and fuse them into a sentence, before generating the final output (see \cref{fig:MDS_CoT_prompt,fig:LFQA_CoT_prompt}). We find that 3 in-context examples and $T=0.1$ work best for MDS, while for LFQA it is 1 example and $T=0.3$ that yield the best results on the development set.

\paragraph{Fine-tuned}
Examples are provided in \cref{fig:ft_MDS_sentence_generation,fig:ft_lfqa_sentence_generation}.
The input to the attributable fusion model is a single cluster of highlights, where highlights are separated by a special markup. The output is then a single sentence of the response. This model is called iteratively, and at each iteration, we include the prefix of the already generated response. The output also includes citations for every sentence in the generated response based on the highlights used to generate the particular sentence. The hyperparameters for the fusion model are for MDS are warmup steps 54, weight decay 0.2, and learning rate 1e-4. The hyperparameters used for LFQA are warmup steps 175, weight decay 0.5, and learning rate 2e-5.

\subsection{End-to-end Generation}\label{subsec:end-to-end-gen}
We include as a baseline end-to-end generation without decomposing the process into several steps.

\paragraph{ICL}
For the ICL setting, we instruct the model to perform the corresponding setting's task, namely summarization for MDS, and answering a query based on the supplied document for LFQA. In addition to the varying number of in-context examples, we also evaluate the model in a zero-shot setting. Our analyses show that for MDS, 1 example and $T=0.5$ give the best performance, while for LFQA 3 examples and $T=0.7$ are optimal.

\paragraph{Fine-tuned}
The input to the model is the set of all documents, concatenated with a special markup ``<doc-sep>'', which was also used in pre-training in \textsc{Primera}. Similar to our content selection, we trim each document to the length of 4096 divided by the number of documents in each example. The output is simply the full response, a summary for MDS, or an answer for LFQA. The hyperparameters used for both MDS and LFQA are warmup steps 27, weight decay 0.2, and learning rate 4e-5.

\subsection{ALCE Variant}\label{subsec:ALCE_variant}
As another baseline, we also implement the ALCE approach \citep{gao-etal-2023-enabling} for attributed generation, where the model is guided to jointly generate a text with in-line citations, pointing to the supporting source documents. We use the same prompt as in the original paper. Additionally, we find that for both settings, 2 examples and $T=0.5$ yield the best results on the respective development sets.

\subsection{Constrained Decoding} \label{appendix:logits_processor}

Both the content selection component and the planning component are designed to exactly copy text spans from source documents. To guide the generation to precisely follow the lexical structure of the input's text, we employ a constrained decoding strategy for the fine-tuned models. This strategy relies on a logits processor\footnote{\url{https://huggingface.co/docs/transformers/internal/generation_utils\#logitsprocessor}}, allowing the model to avoid certain tokens by assigning their scores to minus infinity. Specifically, the model is restricted from generating n-grams absent from the source documents.
Additionally, to optimize the models for producing concise and informative highlights, we impose additional constraints, including setting minimum thresholds on the number of words in each highlight and the overall number of highlights generated. Based on hyperparameter-tuning we set a minimum of three words for each highlight in both MDS and LFQA, and a minimum number of 30 highlights for MDS and 5 highlights for LFQA. These constraints contribute significantly to optimizing the performance and reliability of the fine-tuned generation models.
Lastly, in the planning component, the model organizes and forms clusters, each encapsulating highlights from the given input. To prevent excessively large clusters, we impose a maximum threshold of two highlights per cluster. This facilitates the generation of coherent sentences from each cluster.

\input{tables/tab_autoais_sent}

\section{Full-sentence \textsc{AutoAIS}}\label{sec:full_sentence_autoais}

\cref{tab:autoais_sent_qa,tab:autoais_sent_mds} present the results on the full-sentence AutoAIS vs. regular AutoAIS

\input{tables/tab_example_predictions}

\section{Fine-grained Attribution Examples} \label{sec:fine-grained_attribution_examples}
\cref{tab:example_predictions} shows fine-grained attributions examples, generated by \textsc{Attr. First}\textsubscript{\texttt{CoT}} in the MDS setting.

% \section{Evaluation Details}\label{sec:evaluation_details}

\begin{figure*}[t!]
\begin{adjustbox}{center}
    \centering
    \includegraphics[width=\textwidth]{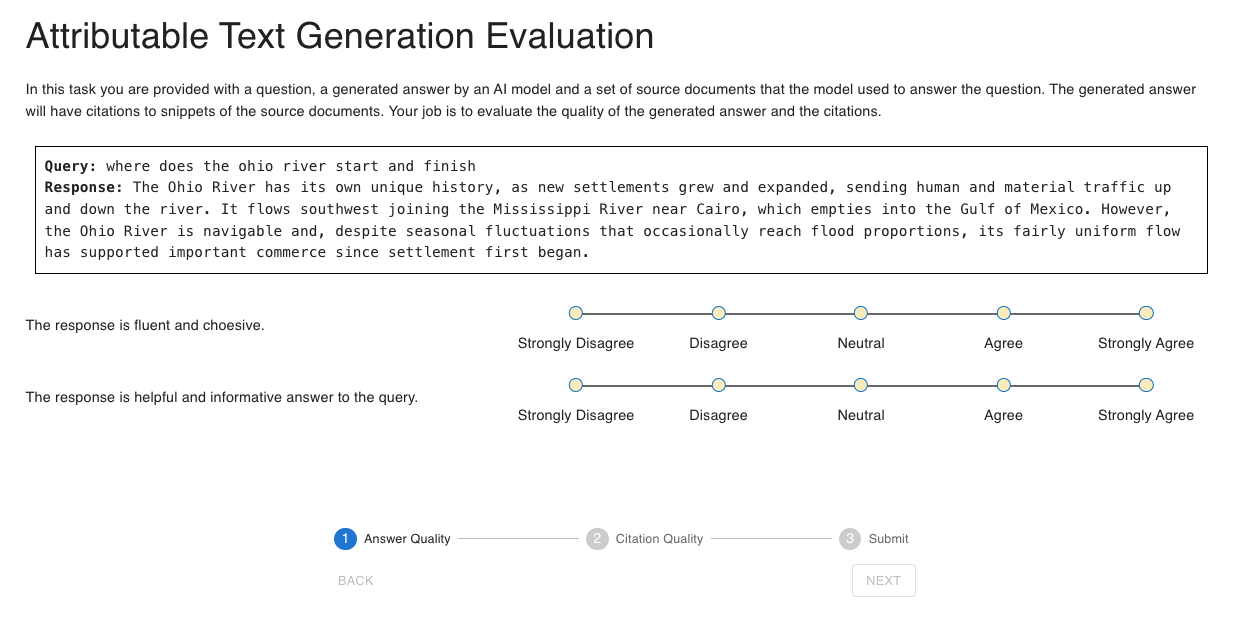}
\end{adjustbox}
    \caption{A screenshot from the first step in our human evaluation annotation interface for the LFQA setting, which is a replication of the interface presented by \citet{liu-etal-2023-evaluating}. The annotator is presented with a query and a response and has to rate two aspects of the response on a Likert scale between 1 to 5. The first aspect is the fluency and cohesiveness of the response. The second is the helpfulness of the response.}
    \label{fig:annotation_step1}
\end{figure*}

\begin{figure*}[t!]
\begin{adjustbox}{center}
    \centering
    \includegraphics[width=\textwidth]{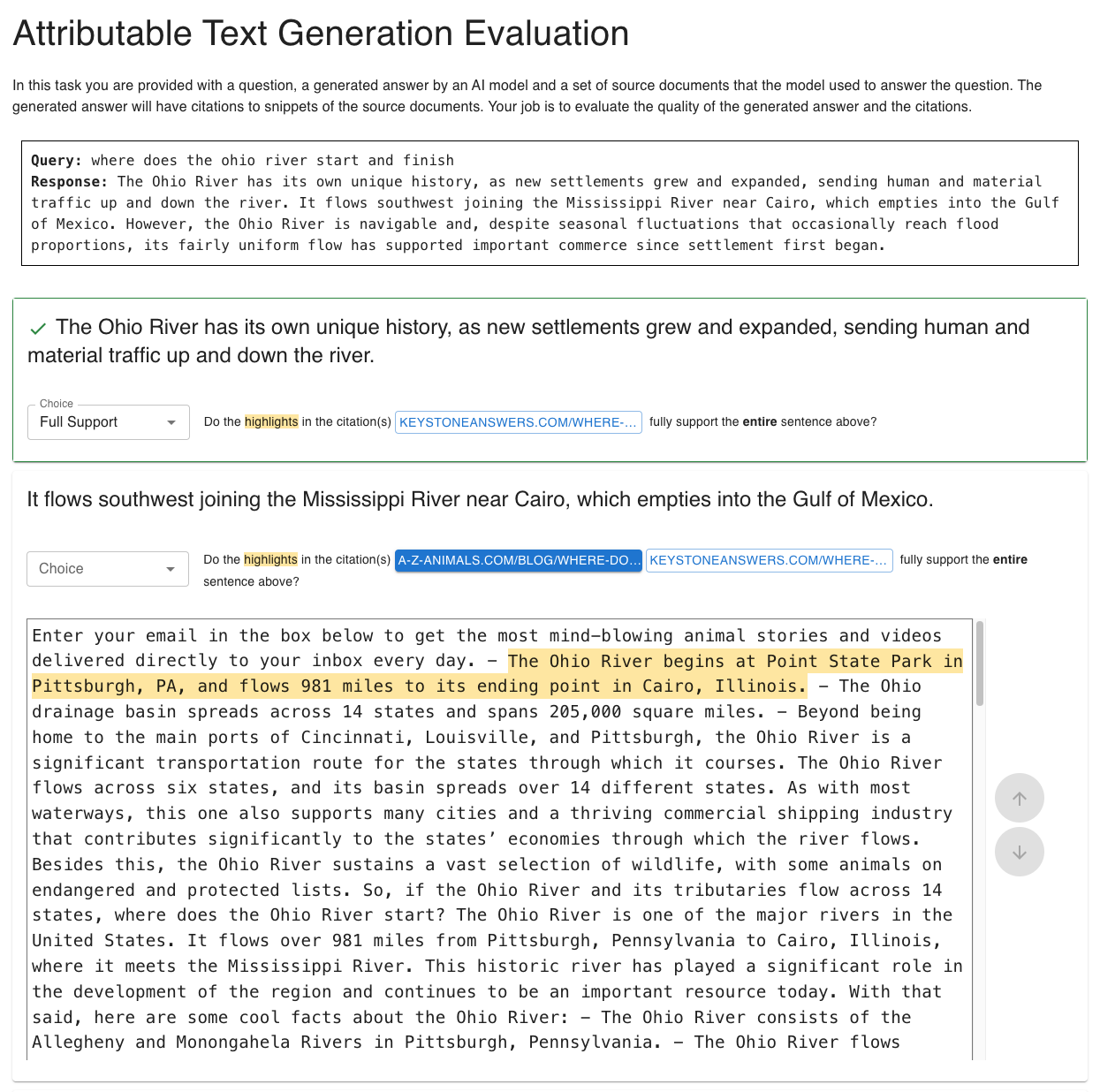}
\end{adjustbox}
    \caption{A screenshot from the second step in our human evaluation annotation interface for the LFQA setting, which is an adaption of the interface suggested by \citet{liu-etal-2023-evaluating}. The annotator is presented with citations provided by the model. For each sentence, they have to rate the quality of the citations. Highlights are presented if available, with arrows to navigate between different highlights. Completed sentences are presented with a green arrow to indicate that the annotator can continue to the next sentence.}
    \label{fig:annotation_step2}
\end{figure*}

% \input{tables/tab_ctr}

% \section{Plan Recovery Algorithm Ablation}\label{sec:backtracking_ablation}
% In \cref{sec:modeling} we explore an algorithm designed to extract a sentence-level planning structure from a surrogate summary produced by an LLM, which is then used to guide the re-generation of the output summary during the sentence fusion phase
% ((c) in \cref{fig:architecture}). 
% To evaluate the effectiveness of the additional re-generation step, we also consider directly using the surrogate summary. Our findings, detailed in \cref{tab:ctr_ablation_mds,tab:ctr_ablation_lfqa}, reveal that, except for one case, this additional step leads to improved output in terms of the actual task, as indicated by the \textsc{Rouge$_L$} and \textsc{BertScore} metrics. Notably, we find that this approach is significantly boosts attribution accuracy, particularly for the ICL variants, where the re-generation step nearly doubles the \textsc{AutoAIS} score.

% These findings suggest that while LLMs excel at generating coherent texts, they face challenges when required to adhere to multiple constraints simultaneously. Therefore, providing additional, more targeted guidance that breaks down these constraints into smaller, more manageable tasks could significantly enhance the model's ability to comply with them effectively.

\section{Human Study}\label{sec:human_study}

% \input{tables/tab_human_evaluation}

% Human evaluation results are reported in \cref{tab:human_results_mds,tab:human_results_qa}.
For our human study, we followed the Controlled Crowdsourcing Methodology \cite{roit-etal-2020-controlled} to detect qualified annotators via Amazon Mechanical Turk.\footnote{\url{www.mturk.com}} 
We reached out to potential candidates, who had exhibited high proficiency in similar tasks in the past, through email invitations sent from the MTurk platform.
During onboarding, candidates reviewed the annotation guidelines and completed a closed qualification round consisting of five annotation tasks. This process, overseen by an author of the paper who provided individualized feedback, required approximately 20 minutes to complete, with participants compensated 4.5\$ for their efforts.
Ultimately, six out of eleven candidates were selected for the actual evaluations, with compensation rates set at 2\$ per ALCE instance and 1\$ otherwise, reflecting the varying complexity of the instances. To ensure fair compensation, we also monitored the actual time required to complete each instance and offered post-hoc reimbursements when this time exceeded our initial estimates by more than 5\%.

Moreover, to evaluate the consistency of attribution annotations, we computed the inter-annotator agreement (IAA) using Fleiss' Kappa coefficient \citep{fleiss1971measuring}, based on a sample of 101 sentences and their attributions from 23 instances across both models, each annotated by 3 workers.\footnote{We treat annotators' scores as binary, akin to \cref{sec:annotation_interface}.}
The resulting Kappa coefficient was 0.37, indicating a fair level of agreement \citep{landis1977measurement}.

\section{Annotation Interface}\label{sec:annotation_interface}
Screenshots from our annotation interface are presented in \cref{fig:annotation_step1,fig:annotation_step2}. Our interface mirrors the design from \citet{liu-etal-2023-evaluating}, adapted to support citations with highlights. In addition, the documents are presented within the interface rather than via external links. 
Within this interface, annotators assess each sentence to determine the level of support provided by its accompanying citations, choosing one of the following options: ``Full Support'', ``Partial Support'', ``No Support'', ``Citation Has Support but also Refutes Statement'', ``Statement is Unclear, Can't Make Judgement''. 
Finally, when calculating the final AIS score, we treat instances labeled as ``Partial Support'' as \textit{non-attributed}, while instances classified as ``Citation Has Support but also Refutes Statement'' are labeled as \textit{attributed}, akin to \citet{gao-etal-2023-rarr}. 

\section{Partially-attributing Examples}\label{sec:partially-attributing-examples}

\cref{tab:partially_attributed_examples} shows a few attribution examples labeled by human annotators as partially-supporting.

\input{tables/tab_partially_attributed_examples}

\input{tables/tab_prefix_examples}

\input{tables/token_count}

\input{tables/tab_latency}

\section{With Prefix Versus without Prefix}\label{sec:prefix_vs_no_prefix}

\cref{tab:prefix_examples} presents several instances from our test set when the iterative sentence-level fusion was only conditioned on each iteration's cluster, versus when it was also conditioned on the prefix.

\section{Latency and Total Number of Input and Output Tokens}
\label{sec:latency_and_tkn_count}
In \cref{tab:token_count} we present the average number of tokens processed in the input and also the average number of generated tokens for each of our pipeline's subtasks, and also for our baselines. Additionally, in \cref{tab:latency} we present the average time in seconds it takes to run each of our pipeline's subtasks, as well as our baselines.

\section{Post-generation Attribution}
\label{sec:post-generation_attribution}
We also investigated a post-generation attribution variant, where first an unattributed output was generated, and then relevant attribution was retrieved. To that end, we first generated the task-specific output, using our baseline from Section~\ref{subsec:models}. 
Then, to obtain attributions, we adapted the RARR algorithm, as outlined by \citet{gao-etal-2023-rarr}, to suit our context. In contrast to its original application of retrieving relevant documents, our adaptation employs the grounding texts directly. The process involves generating queries from each generated sentence about the various facets it encompasses, identifying 4-sentence paragraphs within the source texts that correspond to each query through a query-document relevance model, and answering query based on these paragraphs and the output sentence. 
The comparisons of these responses determine the selection of attributing paragraphs; only those that either entail or contradict are retained as valid attributions.
The results, as detailed in \cref{tab:rarr_results}, show that this method produced both longer and less accurate attributions compared to those derived from our approach.

\input{tables/tab_rarr}

%% file: tables/tab_datasets_sizes.tex
\begin{table}[t!]
    \centering
    \small
    \adjustbox{max width=\columnwidth}{
        \begin{tabular}{lccc}
             Method & \multicolumn{1}{c}{Train} & \multicolumn{1}{c}{Eval} & \multicolumn{1}{c}{Test} \\
             \toprule
             \textsc{Evaluating \citep{liu-etal-2023-evaluating}} & 393 & 44 & 45 \\
             \textsc{Multi-News \citep{fabbri-etal-2019-multi}} & - & 28 & 65 \\
             \textsc{DUC \& TAC (\url{duc.nist.gov})} & 475 & 119 & - \\
             \bottomrule
        \end{tabular}
    }
    \vspace{-4pt}
    \caption{Our different datasets sizes used for training, development, testing and hyperparameter-tuning.}
    \label{tab:datasets_sizes}
    % \vspace{-4pt}
\end{table}

%% file: tables/correlation_with_human_judgement.tex
% \usepackage{graphicx}
\begin{table}[t!]
\centering
\resizebox{\columnwidth}{!}{%
\begin{tabular}{l|cc|cc}
\hline
                       & \multicolumn{2}{c|}{\textbf{Kendall-Tau}} & \multicolumn{2}{c}{\textbf{Spearman}} \\
                       & {\small \textbf{$\tau$}}       & {\small \textbf{CI}}       & {\small \textbf{$\tau$}}     & {\small \textbf{CI}}     \\ \hline
TRUE          & 0.\textbf{4659}                & \textbf{0.46-0.47}         & \textbf{0.5215}              & \textbf{0.51-0.53}       \\ \hline
TRUE\textsubscript{\texttt{prob.}}            & 0.3480                & 0.34-0.36         & 0.4182              & 0.41-0.43       \\ \hline
TrueTeacher & 0.4550                & 0.45-0.46         & 0.5130              & 0.50-0.52       \\ \hline
TrueTeacher\textsubscript{\texttt{prob.}}     & 0.3408                & 0.33-0.35         & 0.4152              & 0.41-0.42       \\ \hline
\end{tabular}%
}
\caption{Average Kendall-Tau and Spearman's rank correlations ($\tau$) and their $95\%$ confidence intervals (CI) for tested evaluation metrics against human judgment. ``prob.'' refers to the variants averaging the respective model's probabilities of the label '1'.}
% \vspace{-0.3cm}
\label{tab:correlation_with_human_judgement}
\end{table}

%% file: resources/MDS_content_selection_prompt.tex
\begin{figure*}[t]

\lstdefinestyle{promptStyle}
{
    basicstyle={\footnotesize\ttfamily},% footnotesize acceptable for monospace
    numbers=left,numberstyle=\footnotesize,
    xleftmargin=2.8em,% show line numbers, remove this entire line if you don't want the numbers.
    xrightmargin=1.5em,
    showstringspaces=false,
      showspaces=false,
        showtabs=false,
    tabsize=2,
    breaklines=true,
        flexiblecolumns=true,
        escapeinside={<@}{@>},
          breakatwhitespace=true
}

\newtcblisting{mylisting}[1]{
  enhanced,
  listing only,
  boxrule=0.8pt,
  sharp corners=downhill,
  top=0mm,
  bottom=0mm,
  left=2mm,
  right=0mm,
  boxsep=0mm,
  colframe=black,
  colback=white,
  listing options={
    style=#1
  }
}

\definecolor{instructionsColor}{rgb}{0.1, 0.5, 0.1}

\begin{mylisting}{promptStyle}
<@\textcolor{instructionsColor}{In this task, you are presented with several documents, which need to be summarized. As an intermediate step, you need to identify salient content within the documents. For each document, copy verbatim the salient spans, and use <SPAN\_DELIM> as a delimiter between each consecutive span. IMPORTANT: The output must be of the format Document [<DOC\_ID>]: <SPAN\_DELIM>-delimited consecutive salient spans. IMPORTANT: Each salient content must be a single consecutive verbatim span from the corresponding passages. IMPORTANT: make sure the total number of copied words (from all documents) is around 200 words, and not more than 900.}@>

<@\color{blue}Document [1]:@> The United States on Wednesday warned US citizens of a possible spike in "terrorist activity" in Colombia in the coming week as the left-wing Revolutionary Armed Forces of Colombia (FARC) mark the one-year anniversary of the death of one of their commanders ... 
<@\color{blue}Document [2]:@> Colombia's hardline government said Friday it is willing to meet with members of the country's main leftist rebel group in an unprecedented offer aimed at freeing dozens of rebel hostages, including three Americans. Before any face-to-face meeting can take place, however, the rebels must agree to swap 15 hostages for 15 rebels jailed on minor charges, said Peace Commissioner Luis Carlos Restrepo ...
...

<@\color{red}Answer:@>
<@\color{blue}Document [1]:@> The United States on Wednesday warned US citizens <SPAN_DELIM> mark the one-year anniversary of the death of one of their commanders <SPAN_DELIM> ...
<@\color{blue}Document [2]:@> Colombia's hardline government said <SPAN_DELIM> it is willing to meet with members of the country's main leftist rebel group <SPAN_DELIM> ...
...

<@\textcolor{instructionsColor}{In this task, you are presented with several documents, which need to be summarized. As an intermediate step, you need to identify salient content within the documents. For each document, copy verbatim the salient spans, and use <SPAN\_DELIM> as a delimiter between each consecutive span. IMPORTANT: The output must be of the format Document [<DOC\_ID>]: <SPAN\_DELIM>-delimited consecutive salient spans. IMPORTANT: Each salient content must be a single consecutive verbatim span from the corresponding passages. IMPORTANT: make sure the total number of copied words (from all documents) is around 200 words, and not more than 900.}@>

<@\color{blue}Document [1]:@> Voters in 11 states will pick their governors tonight, and Republicans appear on track to increase their numbers by at least one, with the potential to extend their hold to more than two-thirds of the nation's top state offices ...
<@\color{blue}Document [2]:@> Eight of the gubernatorial seats up for grabs today are now held by Democrats; three are in Republican hands. Republicans currently hold 29 governorships, Democrats have 20 ...
...

<@\color{red}Answer:@>
\end{mylisting}
% \newtcbinputlisting[caption=Example prompt provided to GPT-4., frame=tlrb, captionpos=b]{figures/prompt.txt}
\caption{Example prompt for the content selection subtask in the MDS setting.}
\label{fig:MDS_content_selection_prompt}
\end{figure*}

%% file: resources/MDS_clustering_prompt.tex
\begin{figure*}[t]

\lstdefinestyle{promptStyle}
{
    basicstyle={\footnotesize\ttfamily},% footnotesize acceptable for monospace
    numbers=left,numberstyle=\footnotesize,
    xleftmargin=2.8em,% show line numbers, remove this entire line if you don't want the numbers.
    xrightmargin=1.5em,
    showstringspaces=false,
      showspaces=false,
        showtabs=false,
    tabsize=2,
    breaklines=true,
        flexiblecolumns=true,
        escapeinside={<@}{@>},
          breakatwhitespace=true
}

\newtcblisting{mylisting}[1]{
  enhanced,
  listing only,
  boxrule=0.8pt,
  sharp corners=downhill,
  top=0mm,
  bottom=0mm,
  left=2mm,
  right=0mm,
  boxsep=0mm,
  colframe=black,
  colback=white,
  listing options={
    style=#1
  }
}

\definecolor{instructionsColor}{rgb}{0.1, 0.5, 0.1}

\begin{mylisting}{promptStyle}
<@\textcolor{instructionsColor}{In this task, you are presented with several passages, where some parts are "highlighted" (namely, there are <highlight\_start> and <highlight\_end> tokens before and after each such span). The goal is to fuse all those highlights into a single summary. As an intermediate step, you need to cluster highlights that can be merged into a sentence (namely, each cluster will be later merged into one sentence). Make sure the clusters are in the same order you would then write the corresponding summary sentences. IMORTANT: make sure there are at least 7 clusters, and no more than 5-6 highlights per cluster. IMPORTANT: The output must be of the format [\{"cluster":[comma-delimited highlights indices]\}]}@>

<@\color{orange}Document [1]:@> <highlight_start>The United States on Wednesday warned US citizens<highlight_end> of a possible spike in "terrorist activity" in Colombia in the coming week as the left-wing Revolutionary Armed Forces of Colombia (FARC) <highlight_start>mark the one-year anniversary of the death of one of their commanders<highlight_end> ... 
<@\color{orange}Document [2]:@> <highlight_start>Colombia's hardline government said<highlight_end> Friday <highlight_start>it is willing to meet with members of the country's main leftist rebel group<highlight_end> in an unprecedented offer aimed at freeing dozens of rebel hostages, including three Americans. Before any face-to-face meeting can take place, however, the rebels must agree to swap 15 hostages for 15 rebels jailed on minor charges, said Peace Commissioner Luis Carlos Restrepo ...
...

<@\color{blue}The highlighted spans are:@>
<@\color{orange}Document [1]:@> 
 1. The United States on Wednesday warned US citizens
 2. mark the one-year anniversary of the death of one of their commanders 
...
<@\color{orange}Document [2]:@> 
 7. Colombia's hardline government said
 8. it is willing to meet with members of the country's main leftist rebel group 
...

<@\color{red}Answer:@> <@\color{blue}The highlighted spans are clustered as follows:@>
[{"cluster":[1,2,5]}, {"cluster":[7,8]} ...]

<@\textcolor{instructionsColor}{In this task, you are presented with several passages, where some parts are "highlighted" (namely, there are <highlight\_start> and <highlight\_end> tokens before and after each such span). The goal is to fuse all those highlights into a single summary. As an intermediate step, you need to cluster highlights that can be merged into a sentence (namely, each cluster will be later merged into one sentence). Make sure the clusters are in the same order you would then write the corresponding summary sentences. IMORTANT: make sure there are at least 7 clusters, and no more than 5-6 highlights per cluster. IMPORTANT: The output must be of the format [\{"cluster":[comma-delimited highlights indices]\}]}@>

<@\color{orange}Document [1]:@> <highlight_start>Voters in 11 states will pick their governors tonight<highlight_end>, and Republicans appear on track to increase their numbers by at least one, with the potential to extend their hold to more than two-thirds of the nation's top state offices ...
<@\color{orange}Document [2]:@> Eight of the gubernatorial seats up for grabs today are now held by Democrats; three are in Republican hands. <highlight_start>Republicans currently hold 29 governorships, Democrats have 20<highlight_end> ...
...

<@\color{blue}The highlighted spans are:@>
<@\color{orange}Document [1]:@> 
 1. Voters in 11 states will pick their governors tonight
...
<@\color{orange}Document [2]:@> 
 5. Republicans currently hold 29 governorships, Democrats have 20
...
<@\color{red}Answer:@> <@\color{blue}The highlighted spans are clustered as follows:@>
\end{mylisting}
% \newtcbinputlisting[caption=Example prompt provided to GPT-4., frame=tlrb, captionpos=b]{figures/prompt.txt}
\caption{Example prompt for the clustering subtask in the MDS setting.}
\label{fig:MDS_clustering_prompt}
\end{figure*}

%% file: resources/MDS_sent_generation_prompt.tex
\begin{figure*}[t]

\lstdefinestyle{promptStyle}
{
    basicstyle={\footnotesize\ttfamily},% footnotesize acceptable for monospace
    numbers=left,numberstyle=\footnotesize,
    xleftmargin=2.8em,% show line numbers, remove this entire line if you don't want the numbers.
    xrightmargin=1.5em,
    showstringspaces=false,
      showspaces=false,
        showtabs=false,
    tabsize=2,
    breaklines=true,
        flexiblecolumns=true,
        escapeinside={<@}{@>},
          breakatwhitespace=true
}

\newtcblisting{mylisting}[1]{
  enhanced,
  listing only,
  boxrule=0.8pt,
  sharp corners=downhill,
  top=0mm,
  bottom=0mm,
  left=2mm,
  right=0mm,
  boxsep=0mm,
  colframe=black,
  colback=white,
  listing options={
    style=#1
  }
}

\definecolor{instructionsColor}{rgb}{0.1, 0.5, 0.1}

\begin{mylisting}{promptStyle}
<@\textcolor{instructionsColor}{In this task, you are presented with several passages, where some parts are "highlighted" (namely, there are <highlight\_start> and <highlight\_end> tokens before and after each such span). You are also presented with a prefix of a paragraph. You job is to generate the next sentence in the paragraph, that covers all and only the "highlighted" spans. Make sure it connects well with the prefix, and that it covers all and only the "highlighted" spans.}@>

<@\color{orange}Document [1]:@> <highlight_start>The United States on Wednesday warned US citizens<highlight_end> of a possible spike in "terrorist activity" in Colombia in the coming week as the left-wing Revolutionary Armed Forces of Colombia (FARC) <highlight_start>mark the one-year anniversary of the death of one of their commanders<highlight_end> ... 
<@\color{orange}Document [2]:@> Colombia's hardline government said Friday it is willing to meet with members of the country's main leftist rebel group in an unprecedented offer aimed at freeing dozens of rebel hostages, including three Americans ...
...

<@\color{blue}Prefix:@> Colombia's strict government agreed to meet with the Country's main leftist rebel group.

<@\color{blue}The highlighted spans are:@>
<@\color{orange}Document [1]:@> 
 1. The United States on Wednesday warned US citizens
 2. mark the one-year anniversary of the death of a terrorist commanders, American citizens recieved a warning from their government
...
<@\color{red}Answer:@> 
<@\color{blue}The next sentence is: @>Following the one-year anniversary of the death of one of the group's commanders, American citizens received a warning from their government on Wednesday.

<@\textcolor{instructionsColor}{In this task, you are presented with several passages, where some parts are "highlighted" (namely, there are <highlight\_start> and <highlight\_end> tokens before and after each such span). You are also presented with a prefix of a paragraph. You job is to generate the next sentence in the paragraph, that covers all and only the "highlighted" spans. Make sure it connects well with the prefix, and that it covers all and only the "highlighted" spans.}@>

<@\color{orange}Document [1]:@> <highlight_start>Voters in 11 states will pick their governors tonight<highlight_end>, and Republicans appear on track to increase their numbers by at least one, with the potential to extend their hold to more than two-thirds of the nation's top state offices ...
<@\color{orange}Document [2]:@> Eight of the gubernatorial seats up for grabs today are now held by Democrats; three are in Republican hands. <highlight_start>Republicans currently hold 29 governorships, Democrats have 20 ...
...

<@\color{blue}Prefix:@> 29 governorships are currently held by Republicans, vs. Democrat's 20.

<@\color{blue}The highlighted spans are:@>
<@\color{orange}Document [1]:@> 
 1. Voters in 11 states will pick their governors tonight
...
<@\color{red}Answer:@> 
<@\color{blue}The next sentence is: @> 
\end{mylisting}
% \newtcbinputlisting[caption=Example prompt provided to GPT-4., frame=tlrb, captionpos=b]{figures/prompt.txt}
\caption{Example prompt for iterative, sentence-by-sentence generation approach in the MDS setting.}
\label{fig:MDS_sent_generation_prompt}
\end{figure*}

%% file: resources/MDS_CoT_prompt.tex
\begin{figure*}[t]

\lstdefinestyle{promptStyle}
{
    basicstyle={\footnotesize\ttfamily},% footnotesize acceptable for monospace
    numbers=left,numberstyle=\footnotesize,
    xleftmargin=2.8em,% show line numbers, remove this entire line if you don't want the numbers.
    xrightmargin=1.5em,
    showstringspaces=false,
      showspaces=false,
        showtabs=false,
    tabsize=2,
    breaklines=true,
        flexiblecolumns=true,
        escapeinside={<@}{@>},
          breakatwhitespace=true
}

\newtcblisting{mylisting}[1]{
  enhanced,
  listing only,
  boxrule=0.8pt,
  sharp corners=downhill,
  top=0mm,
  bottom=0mm,
  left=2mm,
  right=0mm,
  boxsep=0mm,
  colframe=black,
  colback=white,
  listing options={
    style=#1
  }
}

\definecolor{instructionsColor}{rgb}{0.1, 0.5, 0.1}

\begin{mylisting}{promptStyle}
<@\textcolor{instructionsColor}{In this task, you are presented with several passages, where some parts are "highlighted" (namely, there are <highlight\_start> and <highlight\_end> tokens before and after each such span). Your job is to generate a coherent summary that covers all and only the "highlighted" spans. Build the summary iteratively, where each time you cluster highlights to build the next sentence. Make sure ALL highlights are included. IMPORTANT: make sure that the final summary consists of approximately 200 words and at least 7 sentences. IMPORTANT: The clustering steps must be of the form: "Highlights \{comma-delimited highlights indices\} are combined to form sentence \{sentence\_id\}:{sentence\_text}". Additionally, before generating the final summary, say "So the final summary is:".}@>

<@\color{orange}Document [1]:@> <highlight_start>The United States on Wednesday warned US citizens<highlight_end> of a possible spike in "terrorist activity" in Colombia in the coming week as the left-wing Revolutionary Armed Forces of Colombia (FARC) <highlight_start>mark the one-year anniversary of the death of one of their commanders<highlight_end> ... 
<@\color{orange}Document [2]:@> <highlight_start>Colombia's hardline government said<highlight_end> Friday <highlight_start>it is willing to meet with members of the country's main leftist rebel group<highlight_end> in an unprecedented offer aimed at freeing dozens of rebel hostages, including three Americans ...
...
<@\color{blue}The highlighted spans are:@>
<@\color{orange}Document [1]:@> 
 1. The United States on Wednesday warned US citizens
 2. mark the one-year anniversary of the death of one of their commanders
...
<@\color{orange}Document [2]:@> 
 7. Colombia's hardline government said
 8. it is willing to meet with members of the country's main leftist rebel group 
...
<@\color{red}Answer:@> 
<@\color{blue}The highlighted spans are combined as follows:@>
Highlights 7,8 are combined to form sentence 1: Colombia's strict government agreed to meet with the Country's main leftist rebel group.
...
<@\color{blue}So the final summary is:@>  Colombia's strict government agreed to meet with the Country's main leftist rebel group. ...

<@\textcolor{instructionsColor}{In this task, you are presented with several passages, where some parts are "highlighted" (namely, there are <highlight\_start> and <highlight\_end> tokens before and after each such span). Your job is to generate a coherent summary that covers all and only the "highlighted" spans. Build the summary iteratively, where each time you cluster highlights to build the next sentence. Make sure ALL highlights are included. IMPORTANT: make sure that the final summary consists of approximately 200 words and at least 7 sentences. IMPORTANT: The clustering steps must be of the form: "Highlights \{comma-delimited highlights indices\} are combined to form sentence \{sentence\_id\}:{sentence\_text}". Additionally, before generating the final summary, say "So the final summary is:".}@>

<@\color{orange}Document [1]:@> <highlight_start>Voters in 11 states will pick their governors tonight<highlight_end>, and Republicans appear on track to increase their numbers by at least one, with the potential to extend their hold to more than two-thirds of the nation's top state offices ...
<@\color{orange}Document [2]:@> Eight of the gubernatorial seats up for grabs today are now held by Democrats; three are in Republican hands. <highlight_start>Republicans currently hold 29 governorships, Democrats have 20<highlight_end> ...
...
<@\color{blue}The highlighted spans are:@>
<@\color{orange}Document [1]:@> 
 1. Voters in 11 states will pick their governors tonight
...
<@\color{orange}Document [2]:@> 
 5. Republicans currently hold 29 governorships, Democrats have 20
...
<@\color{red}Answer:@> 
<@\color{blue}The highlighted spans are combined as follows:@>
\end{mylisting}
% \newtcbinputlisting[caption=Example prompt provided to GPT-4., frame=tlrb, captionpos=b]{figures/prompt.txt}
\caption{Example prompt for the CoT approach in the MDS setting.}
\label{fig:MDS_CoT_prompt}
\end{figure*}

%% file: resources/LFQA_content_selection_prompt.tex
\begin{figure*}[t]

\lstdefinestyle{promptStyle}
{
    basicstyle={\footnotesize\ttfamily},% footnotesize acceptable for monospace
    numbers=left,numberstyle=\footnotesize,
    xleftmargin=2.8em,% show line numbers, remove this entire line if you don't want the numbers.
    xrightmargin=1.5em,
    showstringspaces=false,
      showspaces=false,
        showtabs=false,
    tabsize=2,
    breaklines=true,
        flexiblecolumns=true,
        escapeinside={<@}{@>},
          breakatwhitespace=true
}

\newtcblisting{mylisting}[1]{
  enhanced,
  listing only,
  boxrule=0.8pt,
  sharp corners=downhill,
  top=0mm,
  bottom=0mm,
  left=2mm,
  right=0mm,
  boxsep=0mm,
  colframe=black,
  colback=white,
  listing options={
    style=#1
  }
}

\definecolor{instructionsColor}{rgb}{0.1, 0.5, 0.1}

\begin{mylisting}{promptStyle}
<@\textcolor{instructionsColor}{In this task, you are presented with a question and several documents. The goal is to answer the question based on the given documents. As an intermediate step, you need to identify the minimal number of spans within the documents that answer the question. For each document, copy verbatim those spans, and use <SPAN\_DELIM> as a delimiter between each consecutive span. IMPORTANT: The output must be of the format Document [<DOC\_ID>]: <SPAN\_DELIM>-delimited consecutive spans. IMPORTANT: Each such span must be a single consecutive verbatim span from the corresponding passages. IMPORTANT: make sure the total number of copied words (from all passages) is around 100 words, and not more than 200.}@>

<@\color{blue}Question:@> What was the color of the carpet at the Oscars this year, and why wasn't it red?

<@\color{blue}Document [1]:@> ... tasked to reimagine the carpet color, opting to break tradition so the talent could better transition from daytime arrivals to the evening, according to the Hollywood Reporter ... 
<@\color{blue}Document [2]:@> ... rolling out the red carpet. The celebrated carpet at the Oscars -- the place where stars show off their most elegant and outrageous fashions -- will be a "champagne" color this year. It will be the first time ...
...

<@\color{red}Answer:@>
<@\color{blue}Document [1]:@> reimagine the carpet color <SPAN_DELIM> so the talent could better transition from daytime arrivals to the evening <SPAN_DELIM> ...
<@\color{blue}Document [2]:@> The celebrated carpet at the Oscars <SPAN_DELIM> will be a "champagne" color this year. <SPAN_DELIM> ...
...

<@\textcolor{instructionsColor}{In this task, you are presented with a question and several documents. The goal is to answer the question based on the given documents. As an intermediate step, you need to identify the minimal number of spans within the documents that answer the question. For each document, copy verbatim those spans, and use <SPAN\_DELIM> as a delimiter between each consecutive span. IMPORTANT: The output must be of the format Document [<DOC\_ID>]: <SPAN\_DELIM>-delimited consecutive spans. IMPORTANT: Each such span must be a single consecutive verbatim span from the corresponding passages. IMPORTANT: make sure the total number of copied words (from all passages) is around 100 words, and not more than 200.}@>

<@\color{blue}Question:@> What is power in international relations?

<@\color{blue}Document [1]:@> ... the following concepts of political power :
- Power as a goal of states or leaders;
- Power as a measure of influence or control over outcomes, events, actors and issues;
- Power as victory in conflict ...
<@\color{blue}Document [2]:@> ... American political scientist Joseph Nye Jr., who defined power as "the ability to influence the behaviors of others to get the desired outcome." ...
...

<@\color{red}Answer:@>
\end{mylisting}
% \newtcbinputlisting[caption=Example prompt provided to GPT-4., frame=tlrb, captionpos=b]{figures/prompt.txt}
\caption{Example prompt for the content selection subtask in the LFQA setting.}
\label{fig:LFQA_content_selection_prompt}
\end{figure*}

%% file: resources/LFQA_clustering_prompt.tex
\begin{figure*}[t]

\lstdefinestyle{promptStyle}
{
    basicstyle={\footnotesize\ttfamily},% footnotesize acceptable for monospace
    numbers=left,numberstyle=\footnotesize,
    xleftmargin=2.8em,% show line numbers, remove this entire line if you don't want the numbers.
    xrightmargin=1.5em,
    showstringspaces=false,
      showspaces=false,
        showtabs=false,
    tabsize=2,
    breaklines=true,
        flexiblecolumns=true,
        escapeinside={<@}{@>},
          breakatwhitespace=true
}

\newtcblisting{mylisting}[1]{
  enhanced,
  listing only,
  boxrule=0.8pt,
  sharp corners=downhill,
  top=0mm,
  bottom=0mm,
  left=2mm,
  right=0mm,
  boxsep=0mm,
  colframe=black,
  colback=white,
  listing options={
    style=#1
  }
}

\definecolor{instructionsColor}{rgb}{0.1, 0.5, 0.1}

\begin{mylisting}{promptStyle}
<@\textcolor{instructionsColor}{In this task, you are presented with a question and several passages, where some parts are "highlighted" (namely, there are <highlight\_start> and <highlight\_end> tokens before and after each such span). Those spans are supposed to have the information needed to answer the question, and the goal is to fuse all those highlights into a single paragraph that answers the question. As an intermediate step, you need to cluster highlights that can be merged into a sentence (namely, each cluster will be later merged into one sentence). Make sure the clusters are in the same order you would then write the corresponding output paragraph. IMPORTANT: make sure there are at least two clusters. IMPORTANT: The output must be of the format [\{"cluster":[comma-delimited highlights indices]\}]}@>

<@\color{orange}Question:@> What was the color of the carpet at the Oscars this year, and why wasn't it red?

<@\color{orange}Document [1]:@> ... tasked to <highlight_start>reimagine the carpet color<highlight_end>, opting to break tradition <highlight_start>so the talent could better transition from daytime arrivals to the evening<highlight_end>, according to the Hollywood Reporter ... 
<@\color{orange}Document [2]:@> ... rolling out the red carpet. <highlight_start>The celebrated carpet at the Oscars<highlight_end> -- the place where stars show off their most elegant and outrageous fashions -- <highlight_start>will be a "champagne" color this year.<highlight_end> It will be the first time ...
...

<@\color{blue}The highlighted spans are:@>
<@\color{orange}Document [1]:@> 
 1. reimagine the carpet color
 2. so the talent could better transition from daytime arrivals to the evening 
...
<@\color{orange}Document [2]:@> 
 5. The celebrated carpet at the Oscars
 6. will be a "champagne" color this year.
...

<@\color{red}Answer:@> <@\color{blue}The highlighted spans are clustered as follows:@>
[{"cluster":[1,2,5,6]}, {"cluster":[9,11]} ...]

<@\textcolor{instructionsColor}{In this task, you are presented with a question and several passages, where some parts are "highlighted" (namely, there are <highlight\_start> and <highlight\_end> tokens before and after each such span). Those spans are supposed to have the information needed to answer the question, and the goal is to fuse all those highlights into a single paragraph that answers the question. As an intermediate step, you need to cluster highlights that can be merged into a sentence (namely, each cluster will be later merged into one sentence). Make sure the clusters are in the same order you would then write the corresponding output paragraph. IMPORTANT: make sure there are at least two clusters. IMPORTANT: The output must be of the format [\{"cluster":[comma-delimited highlights indices]\}]}@>

<@\color{orange}Question:@> What is power in international relations?

<@\color{orange}Document [1]:@> ... <highlight_start>Power as a measure of influence or control over outcomes, events, actors and issues;<highlight_end>
- <highlight_start>Power as victory in conflict<highlight_end> ...
<@\color{orange}Document [2]:@> ... American political scientist Joseph Nye Jr., who defined power as <highlight_start>"the ability to influence the behaviors of others to get the desired outcome."<highlight_end> ...
...

<@\color{blue}The highlighted spans are:@>
<@\color{orange}Document [1]:@> 
 1. Power as a measure of influence or control over outcomes, events, actors and issues;
...
<@\color{orange}Document [2]:@> 
 5. "the ability to influence the behaviors of others to get the desired outcome."
...
<@\color{red}Answer:@> <@\color{blue}The highlighted spans are clustered as follows:@>
\end{mylisting}
% \newtcbinputlisting[caption=Example prompt provided to GPT-4., frame=tlrb, captionpos=b]{figures/prompt.txt}
\caption{Example prompt for the clustering subtask in the LFQA setting.}
\label{fig:LFQA_clustering_prompt}
\end{figure*}

%% file: resources/LFQA_sent_generation_prompt.tex
\begin{figure*}[t]

\lstdefinestyle{promptStyle}
{
    basicstyle={\footnotesize\ttfamily},% footnotesize acceptable for monospace
    numbers=left,numberstyle=\footnotesize,
    xleftmargin=2.8em,% show line numbers, remove this entire line if you don't want the numbers.
    xrightmargin=1.5em,
    showstringspaces=false,
      showspaces=false,
        showtabs=false,
    tabsize=2,
    breaklines=true,
        flexiblecolumns=true,
        escapeinside={<@}{@>},
          breakatwhitespace=true
}

\newtcblisting{mylisting}[1]{
  enhanced,
  listing only,
  boxrule=0.8pt,
  sharp corners=downhill,
  top=0mm,
  bottom=0mm,
  left=2mm,
  right=0mm,
  boxsep=0mm,
  colframe=black,
  colback=white,
  listing options={
    style=#1
  }
}

\definecolor{instructionsColor}{rgb}{0.1, 0.5, 0.1}

\begin{mylisting}{promptStyle}
<@\textcolor{instructionsColor}{In this task, you are presented with a question and several passages, where some parts are "highlighted" (namely, there are <highlight\_start> and <highlight\_end> tokens before and after each such span). You are also presented with a prefix of a paragraph. The goal is to generate a paragraph that answers the question, based on the passages. The given prefix is what you generated so far. You job is to generate the next sentence in the paragraph, that covers all and only the "highlighted" spans. Make sure it connects well with the prefix, that it covers all and only the "highlighted" spans, and that it properly builds towards a full and coherent response to the question.}@>

<@\color{orange}Question:@> What was the color of the carpet at the Oscars this year, and why wasn't it red?

<@\color{orange}Document [1]:@> ... interview published Friday."<highlight_start>That's always been something that the Oscars has had a problem with ever since it's started because it begins so early in the day with the sunshine and the heat<highlight_end>," she added. "<highlight_start>But everybody's dressed up for a night event<highlight_end> ... 
...

<@\color{blue}Prefix:@> The Oscars red carpet was changed to a champagne color this year, to better transition from daytime to evening arrivals.

<@\color{blue}The highlighted spans are:@>
<@\color{orange}Document [1]:@> 
 1. That's always been something that the Oscars has had a problem with ever since it's started because it begins so early in the day with the sunshine and the heat
 2. But everybody's dressed up for a night event
...
<@\color{red}Answer:@> 
<@\color{blue}The next sentence is: @>This transition was necessary, as the Oscars always began very early in the day, when it was hot, but everybody was dressed for a night event.

<@\textcolor{instructionsColor}{In this task, you are presented with a question and several passages, where some parts are "highlighted" (namely, there are <highlight\_start> and <highlight\_end> tokens before and after each such span). You are also presented with a prefix of a paragraph. The goal is to generate a paragraph that answers the question, based on the passages. The given prefix is what you generated so far. You job is to generate the next sentence in the paragraph, that covers all and only the "highlighted" spans. Make sure it connects well with the prefix, that it covers all and only the "highlighted" spans, and that it properly builds towards a full and coherent response to the question.}@>

<@\color{orange}Document [1]:@> ... the following concepts of political power :
- Power as a goal of states or leaders;
- <highlight_start>Power as a measure of influence or control over outcomes, events, actors and issues<highlight_end> ... 
...

<@\color{blue}Prefix:@> Power in international relations can be defined as the ability to influence or control outcomes, events, actors, and issues.

<@\color{blue}The highlighted spans are:@>
<@\color{orange}Document [1]:@> 
 1. Power as a measure of influence or control over outcomes, events, actors and issues
...
<@\color{red}Answer:@> 
<@\color{blue}The next sentence is: @> 
\end{mylisting}
% \newtcbinputlisting[caption=Example prompt provided to GPT-4., frame=tlrb, captionpos=b]{figures/prompt.txt}
\caption{Example prompt for iterative, sentence-by-sentence generation approach in the LFQA setting.}
\label{fig:LFQA_sent_generation_prompt}
\end{figure*}

%% file: resources/LFQA_CoT_prompt.tex
\begin{figure*}[t]

\lstdefinestyle{promptStyle}
{
    basicstyle={\footnotesize\ttfamily},% footnotesize acceptable for monospace
    numbers=left,numberstyle=\footnotesize,
    xleftmargin=2.8em,% show line numbers, remove this entire line if you don't want the numbers.
    xrightmargin=1.5em,
    showstringspaces=false,
      showspaces=false,
        showtabs=false,
    tabsize=2,
    breaklines=true,
        flexiblecolumns=true,
        escapeinside={<@}{@>},
          breakatwhitespace=true
}

\newtcblisting{mylisting}[1]{
  enhanced,
  listing only,
  boxrule=0.8pt,
  sharp corners=downhill,
  top=0mm,
  bottom=0mm,
  left=2mm,
  right=0mm,
  boxsep=0mm,
  colframe=black,
  colback=white,
  listing options={
    style=#1
  }
}

\definecolor{instructionsColor}{rgb}{0.1, 0.5, 0.1}

\begin{mylisting}{promptStyle}
<@\textcolor{instructionsColor}{In this task, you are presented with a question and several passages, where some parts are "highlighted" (namely, there are <highlight\_start> and <highlight\_end> tokens before and after each such span). Those spans are supposed to have the information needed to answer the question. Your job is to generate a coherent paragraph that answers the question, while covering all and only the "highlighted" spans. Build this paragraph iteratively, where each time you cluster highlights to build the next sentence. Make sure all highlights are included, and that the output paragraph has at least two sentences. IMPORTANT: The clustering steps must be of the form: "Highlights {comma-delimited highlights indices} are combined to form sentence {sentence\_id}:{sentence\_text}". Additionally, before generating the final output paragraph, say "So the final answer is:".}@>

<@\color{orange}Question:@> What was the color of the carpet at the Oscars this year, and why wasn't it red?

<@\color{orange}Document [1]:@> ... tasked to <highlight_start>reimagine the carpet color<highlight_end>, opting to break tradition <highlight_start>so the talent could better transition from daytime arrivals to the evening<highlight_end>, according to ... 
<@\color{orange}Document [2]:@> ... <highlight_start>The celebrated carpet at the Oscars<highlight_end> -- the place where stars show off their most elegant and outrageous fashions -- <highlight_start>will be a "champagne" color this year.<highlight_end> It will be ...
...

<@\color{red}Answer:@> 
<@\color{blue}The highlighted spans are:@>
<@\color{orange}Document [1]:@> 
 1. reimagine the carpet color
 2. so the talent could better transition from daytime arrivals to the evening 
...
<@\color{orange}Document [2]:@> 
 5. The celebrated carpet at the Oscars
 6. will be a "champagne" color this year.
...
<@\color{blue}The highlighted spans are combined as follows:@>
Highlights 1,2,5,6 are combined to form sentence 1: The Oscars red carpet was changed to a champagne color this year, to better transition from daytime to evening arrivals.
...
<@\color{blue}So the final summary is:@> The Oscars red carpet was changed to a champagne color this year, to better transition from daytime to evening arrivals. ...

<@\textcolor{instructionsColor}{In this task, you are presented with a question and several passages, where some parts are "highlighted" (namely, there are <highlight\_start> and <highlight\_end> tokens before and after each such span). Those spans are supposed to have the information needed to answer the question. Your job is to generate a coherent paragraph that answers the question, while covering all and only the "highlighted" spans. Build this paragraph iteratively, where each time you cluster highlights to build the next sentence. Make sure all highlights are included, and that the output paragraph has at least two sentences. IMPORTANT: The clustering steps must be of the form: "Highlights {comma-delimited highlights indices} are combined to form sentence {sentence\_id}:{sentence\_text}". Additionally, before generating the final output paragraph, say "So the final answer is:".}@>

<@\color{orange}Question:@> What is power in international relations?

<@\color{orange}Document [1]:@> ... <highlight_start>Power as a measure of influence or control over outcomes, events, actors and issues;<highlight_end> Power as victory in conflict ...
<@\color{orange}Document [2]:@> ... Joseph Nye Jr., who defined power as <highlight_start>"the ability to influence the behaviors of others to get the desired outcome."<highlight_end> ...
...

<@\color{red}Answer:@>
<@\color{blue}The highlighted spans are:@>
<@\color{orange}Document [1]:@> 
 1. Power as a measure of influence or control over outcomes, events, actors and issues;
...
<@\color{orange}Document [2]:@> 
 5. "the ability to influence the behaviors of others to get the desired outcome."
...
<@\color{blue}The highlighted spans are combined as follows:@>
\end{mylisting}
% \newtcbinputlisting[caption=Example prompt provided to GPT-4., frame=tlrb, captionpos=b]{figures/prompt.txt}
\vspace{-4pt}
\caption{Example prompt for the CoT approach in the LFQA setting.}
\label{fig:LFQA_CoT_prompt}
\end{figure*}

%% file: resources/FT_MDS_examples_io.tex
\begin{figure*}[t]

\lstdefinestyle{promptStyle}
{
    basicstyle={\footnotesize\ttfamily},% footnotesize acceptable for monospace
    numbers=left,numberstyle=\footnotesize,
    xleftmargin=2.8em,% show line numbers, remove this entire line if you don't want the numbers.
    xrightmargin=1.5em,
    showstringspaces=false,
      showspaces=false,
        showtabs=false,
    tabsize=2,
    breaklines=true,
        flexiblecolumns=true,
        escapeinside={<@}{@>},
          breakatwhitespace=true
}

\newtcblisting{mylisting}[1]{
  enhanced,
  listing only,
  boxrule=0.8pt,
  sharp corners=downhill,
  top=0mm,
  bottom=0mm,
  left=2mm,
  right=0mm,
  boxsep=0mm,
  colframe=black,
  colback=white,
  listing options={
    style=#1
  }
}

\definecolor{instructionsColor}{rgb}{0.1, 0.5, 0.1}

\begin{mylisting}{promptStyle}
<@\color{red}Input:@>
Those documents were cleared earlier Thursday morning, according to spokesmen for both sides. Nevertheless, the drama over the emails once marked ""committee confidential"" engulfed much of the third day of Kavanaugh's confirmation hearings, with Booker vowing to disclose them in defiance of Senate rules.

""I openly invite and accept the consequences of releasing that email right now,"" Booker said. ""

Such theatrics have characterized Kavanaugh's hearings, in which Democrats have repeatedly complained that Republicans have withheld documents from the committee and the public that shed important light on Kavanaugh's past.
...
<@\color{orange}<doc-sep>@>
New Jersey Democrat Cory Booker said he was ready to risk expulsion from the Senate for making public documents pertinent Kavanaugh's time as a top White House aide to President George W. Bush, but the GOP mocked him for grandstanding ahead of a possible 2020 run -- saying it had already cleared the emails for release.

In a striking political gambit, Booker, backed up by Democratic Sen. Mazie Hirono of Hawaii, released emails -- which had been designated as ""committee confidential"" -- that reference Kavanaugh's position on racial profiling and thoughts on Roe v. Wade dating from his time as a White House official under Bush.
...

<@\color{red}Output:@> 
the drama over the emails once marked ""committee confidential"" engulfed much of the third day of Kavanaugh's confirmation hearings, with Booker vowing to disclose them in defiance of Senate rules<@\color{blue}<highlight\_sep>@>""I openly invite and accept the consequences of releasing that email right now,"" Booker said<@\color{blue}<highlight\_sep>@>Democrats have repeatedly complained that Republicans have withheld documents from the committee and the public that shed important light on Kavanaugh's past...<@\color{orange}<doc-sep>@>Booker<@\color{blue}<highlight\_sep>@>released emails -- which had been designated as ""committee confidential"" -- that reference Kavanaugh's position...

\end{mylisting}
% \newtcbinputlisting[caption=Example prompt provided to GPT-4., frame=tlrb, captionpos=b]{figures/prompt.txt}
\caption{Example input and output for the fine-tuned content selection model in the MDS setting. Highlight start and end markups are colored blue.}
\label{fig:ft_MDS_content_selection}
\end{figure*}

\begin{figure*}[t]

\lstdefinestyle{promptStyle}
{
    basicstyle={\footnotesize\ttfamily},% footnotesize acceptable for monospace
    numbers=left,numberstyle=\footnotesize,
    xleftmargin=2.8em,% show line numbers, remove this entire line if you don't want the numbers.
    xrightmargin=1.5em,
    showstringspaces=false,
      showspaces=false,
        showtabs=false,
    tabsize=2,
    breaklines=true,
        flexiblecolumns=true,
        escapeinside={<@}{@>},
          breakatwhitespace=true
}

\newtcblisting{mylisting}[1]{
  enhanced,
  listing only,
  boxrule=0.8pt,
  sharp corners=downhill,
  top=0mm,
  bottom=0mm,
  left=2mm,
  right=0mm,
  boxsep=0mm,
  colframe=black,
  colback=white,
  listing options={
    style=#1
  }
}

\definecolor{instructionsColor}{rgb}{0.1, 0.5, 0.1}

\begin{mylisting}{promptStyle}
<@\color{red}Input:@> 
Those documents were cleared earlier Thursday morning, according to spokesmen for both sides. Nevertheless, <@\color{blue}<highlight\_start>@>the drama over the emails once marked ""committee confidential"" engulfed much of the third day of Kavanaugh's confirmation hearings, with Booker vowing to disclose them in defiance of Senate rules<@\color{blue}<highlight\_end>@>.

<@\color{blue}<highlight\_start>@>""I openly invite and accept the consequences of releasing that email right now,"" Booker said<@\color{blue}<highlight\_end>@>. ""

Such theatrics have characterized Kavanaugh's hearings, in which <@\color{blue}<highlight\_start>@>Democrats have repeatedly complained that Republicans have withheld documents from the committee and the public that shed important light on Kavanaugh's past<@\color{blue}<highlight\_end>@>.
...
<@\color{orange}<doc-sep>@>
New Jersey Democrat Cory Booker said he was ready to risk expulsion from the Senate for making public documents pertinent Kavanaugh's time as a top White House aide to President George W. Bush, but the GOP mocked him for grandstanding ahead of a possible 2020 run -- saying it had already cleared the emails for release.

In a striking political gambit, <@\color{blue}<highlight\_start>@>Booker<@\color{blue}<highlight\_end>@>, backed up by Democratic Sen. Mazie Hirono of Hawaii, <@\color{blue}<highlight\_start>@>released emails -- which had been designated as ""committee confidential"" -- that reference Kavanaugh's position<@\color{blue}<highlight\_end>@> on racial profiling and thoughts on Roe v. Wade dating from his time as a White House official under Bush.
...

<@\color{red}Output:@> 
the drama over the emails once marked ""committee confidential"" engulfed much of the third day of Kavanaugh's confirmation hearings, with Booker vowing to disclose them in defiance of Senate rules<@\color{orange}<doc-sep>@>Booker<@\color{blue}<highlight\_separator>@>released emails -- which had been designated as ""committee confidential"" -- that reference Kavanaugh's position<@\color{instructionsColor}<cluster\_separator>@>""I openly invite and accept the consequences of releasing that email right now,"" Booker said...

\end{mylisting}
% \newtcbinputlisting[caption=Example prompt provided to GPT-4., frame=tlrb, captionpos=b]{figures/prompt.txt}
\caption{Example input and output for the fine-tuned sentence planning model in the MDS setting. Highlights separators are colored blue and cluster separators are colored green.}
\label{fig:ft_MDS_planning}
\end{figure*}

\begin{figure*}[t]

\lstdefinestyle{promptStyle}
{
    basicstyle={\footnotesize\ttfamily},% footnotesize acceptable for monospace
    numbers=left,numberstyle=\footnotesize,
    xleftmargin=2.8em,% show line numbers, remove this entire line if you don't want the numbers.
    xrightmargin=1.5em,
    showstringspaces=false,
      showspaces=false,
        showtabs=false,
    tabsize=2,
    breaklines=true,
        flexiblecolumns=true,
        escapeinside={<@}{@>},
          breakatwhitespace=true
}

\newtcblisting{mylisting}[1]{
  enhanced,
  listing only,
  boxrule=0.8pt,
  sharp corners=downhill,
  top=0mm,
  bottom=0mm,
  left=2mm,
  right=0mm,
  boxsep=0mm,
  colframe=black,
  colback=white,
  listing options={
    style=#1
  }
}

\definecolor{instructionsColor}{rgb}{0.1, 0.5, 0.1}

\begin{mylisting}{promptStyle}
<@\color{red}Input:@> 
Those documents were cleared earlier Thursday morning, according to spokesmen for both sides. Nevertheless, <@\color{blue}<highlight\_start>@>the drama over the emails once marked "committee confidential" engulfed much of the third day of Kavanaugh\'s confirmation hearings, with Booker vowing to disclose them in defiance of Senate rules<@\color{blue}<highlight\_end>@>. <@\color{orange}<doc-sep>@>New Jersey Democrat Cory Booker said he was ready to risk expulsion from the Senate for making public documents pertinent Kavanaugh\'s time as a top White House aide to President George W. Bush, but the GOP mocked him for grandstanding ahead of a possible 2020 run -- saying it had already cleared the emails for release. In a striking political gambit, <@\color{blue}<highlight\_start>@>Booker<@\color{blue}<highlight\_end>@>, backed up by Democratic Sen. Mazie Hirono of Hawaii, <@\color{blue}<highlight\_start>@>released emails -- which had been designated as "committee confidential" -- that reference Kavanaugh\'s position<@\color{blue}<highlight\_end>@> on racial profiling and thoughts on Roe v. Wade dating from his time as a White House official under Bush. The New York Times first reported the email. <@\color{blue}<highlight\_start>@>Throughout the contentious hearing, now in its third day, Kavanaugh has tried to give political questions --<@\color{blue}<highlight\_end>@> especially those related to Trump and his potential legal woes -- a wide berth. <@\color{orange}<doc-sep>@> Senators Dispute Status Of Released Documents  As <@\color{blue}<highlight\_start>@>Questioning Of Kavanaugh Wraps Up<@\color{blue}<highlight\_end>@> Enlarge this image toggle caption Chip Somodevilla/Getty Images Chip Somodevilla/Getty Images Last night, he was admonished by Republicans for breaking the rules when he read from committee confidential documents.  Cory and Senate Democrats were able to shame the committee into agreeing to make last night\'s documents publicly available, and <@\color{blue}<highlight\_start>@>Cory publicly released those documents as well as other committee confidential documents<@\color{blue}<highlight\_end>@> today.  <@\color{blue}<highlight\_start>@>And he\'ll keep releasing them because Republicans are hiding Brett Kavanaugh\'s record from the American people<@\color{blue}<highlight\_end>@>.

<@\color{red}Output:@> 
Day three of the Brett Kavanaugh confirmation hearing has opened with a bang: Democratic Sen. Cory Booker says he has ordered his staff to release a Kavanaugh email even though it's supposed to remain under wraps.

\end{mylisting}
% \newtcbinputlisting[caption=Example prompt provided to GPT-4., frame=tlrb, captionpos=b]{figures/prompt.txt}
\caption{Example input and output for the fine-tuned sentence-by-sentence generation in the MDS setting. This is the first sentence generated in the summary, so there is no prefix.}
\label{fig:ft_MDS_sentence_generation}
\end{figure*}

%% file: resources/FT_LFQA_examples_io.tex
\begin{figure*}[t]

\lstdefinestyle{promptStyle}
{
    basicstyle={\footnotesize\ttfamily},% footnotesize acceptable for monospace
    numbers=left,numberstyle=\footnotesize,
    xleftmargin=2.8em,% show line numbers, remove this entire line if you don't want the numbers.
    xrightmargin=1.5em,
    showstringspaces=false,
      showspaces=false,
        showtabs=false,
    tabsize=2,
    breaklines=true,
        flexiblecolumns=true,
        escapeinside={<@}{@>},
          breakatwhitespace=true
}

\newtcblisting{mylisting}[1]{
  enhanced,
  listing only,
  boxrule=0.8pt,
  sharp corners=downhill,
  top=0mm,
  bottom=0mm,
  left=2mm,
  right=0mm,
  boxsep=0mm,
  colframe=black,
  colback=white,
  listing options={
    style=#1
  }
}

\definecolor{instructionsColor}{rgb}{0.1, 0.5, 0.1}

\begin{mylisting}{promptStyle}

<@\color{red}Input:@> 
Q: 'Wild law': if animals have rights, can nature have rights too?
<@\color{orange}<query\_documents\_separator>@>
The fields of animal law and environmental law have an uneasy relationship. At a basic level, they are intertwined by the fundamental observation that animals, human and nonhuman, exist in the environment. Environmental law is generally concerned with animals at the level of species (and specifically endangered or threatened species), whereas animal law is concerned with all animals, regardless of particular characteristics. The issue of wild horses in the western United States illustrates this tension. Some environmentalists view the horses as "feral pests" that damage the fragile ecosystem and compete with wildlife - and privately owned cattle - for resources.\n
...
<@\color{orange}<doc-sep>@>
Every now and then, a concept emerges with the power to alter orthodoxies. In the legal world, one such concept is the "rights of nature," an idea that can-and should-encompass new protections for wild animals. Every now and then, a concept emerges with the power to alter orthodoxies. In the legal world, one such concept is the "rights of nature." Whereas environmental law traditionally begins from the premise that humans need a healthy environment for their own purposes ( e.g., to enjoy fresh air, clean water, and the many benefits of biodiversity), the rights of nature concept flows from the notion that nature is itself entitled to rights regardless of human interests. The shift is significant: nature moves from being an object of the law to being a subject of the law and a holder of its own rights. In contrast to environmental rights, it is only a fairly recent phenomenon that assigns "constitutional significance to the experiences of individual nonhuman animals."

<@\color{red}Output:@> 
In contrast to environmental rights, it is only a fairly recent phenomenon that assigns "constitutional significance to the experiences of individual nonhuman animals."<@\color{blue}<highlight\_sep>@>At the constitutional level, Ecuador was the first to recognize the rights of nature.<@\color{blue}<highlight\_sep>@>At the most fundamental level, if nature has rights, and if nature includes animals, then rights-based claims could be made on behalf of animals using existing rights of nature doctrine and strategy.<@\color{orange}<doc-sep>@>In the legal world, one such concept is the "rights of nature," an idea that can-and should-encompass new protections for wild animals.<@\color{orange}<doc-sep>@>According to the "Rights of Nature" doctrine, an ecosystem is entitled to legal personhood status and as such, has the right to defend itself in a court of law against harms, including environmental degradation caused by a specific development project or even by climate change.<@\color{orange}<doc-sep>@>Looking at legislation, the recognition of animal sentience is another emerging trend, although this does not go so far as to establish rights for animals.<@\color{blue}<highlight\_sep>@>To assert these rights, giving rivers legal personhood helps protect them as they can then be represented in court.<@\color{blue}<highlight\_sep>@>This has occurred with all the rivers in Bangladesh and the Atrato River in Columbia, and many others.<@\color{orange}<doc-sep>@>In 1972, Supreme Court Justice William O. Douglas first asserted the legal rights of nature in a U.S. court in a dissenting opinion on the Sierra Club v. Morton case.<@\color{blue}<highlight\_sep>@>Earlier this year, the U.S. District Court for the Southern District of Ohio granted the application , thereby officially recognizing animals - hippos - have legal rights akin to humans for the first time in U.S. history.

\end{mylisting}
% \newtcbinputlisting[caption=Example prompt provided to GPT-4., frame=tlrb, captionpos=b]{figures/prompt.txt}
\caption{Example input and output for the fine-tuned content selection model in the LFQA setting. Highlight start and end markups are colored blue.}
\label{fig:ft_lfqa_content_selection}
\end{figure*}

\begin{figure*}[t]

\lstdefinestyle{promptStyle}
{
    basicstyle={\footnotesize\ttfamily},% footnotesize acceptable for monospace
    numbers=left,numberstyle=\footnotesize,
    xleftmargin=2.8em,% show line numbers, remove this entire line if you don't want the numbers.
    xrightmargin=1.5em,
    showstringspaces=false,
      showspaces=false,
        showtabs=false,
    tabsize=2,
    breaklines=true,
        flexiblecolumns=true,
        escapeinside={<@}{@>},
          breakatwhitespace=true
}

\newtcblisting{mylisting}[1]{
  enhanced,
  listing only,
  boxrule=0.8pt,
  sharp corners=downhill,
  top=0mm,
  bottom=0mm,
  left=2mm,
  right=0mm,
  boxsep=0mm,
  colframe=black,
  colback=white,
  listing options={
    style=#1
  }
}

\definecolor{instructionsColor}{rgb}{0.1, 0.5, 0.1}

\begin{mylisting}{promptStyle}
<@\color{red}Input:@> 
Q: 'Wild law': if animals have rights, can nature have rights too? 
<@\color{orange}<query\_documents\_separator>@>
Most constitutions address the environment, and the typical phrasing is anthropocentric: a human right to a healthy environment as seen, for example, in article 42 of the Constitution of Kenya: "Every person has the right to a clean and healthy environment . . . ." Newer trends adopt ecocentric or biocentric approaches and grant rights to nature (or its component parts, such as a river) at the constitutional or legislative level or through judicial decisions. <@\color{blue}<highlight\_start>@>In contrast to environmental rights, it is only a fairly recent phenomenon that assigns "constitutional significance to the experiences of individual nonhuman animals."<@\color{blue}<highlight\_end>@> Ecocentric or biocentric approaches that lodge a right in nature or its component parts also may be promising for the development of legally recognized animal rights. Rights of nature are not widespread, but they have potential for growth and impact. <@\color{blue}<highlight\_start>@>At the constitutional level, Ecuador was the first to recognize the rights of nature.<@\color{blue}<highlight\_end>@> Bolivia adopted this approach through the Law on the Rights of Mother Earth (2010); the enumerated rights are the rights to life, diversity of life, water, clean air, equilibrium, restoration, and pollution-free living. Other countries have recognized the right in judicial opinions. <@\color{blue}<highlight\_start>@>At the most fundamental level, if nature has rights, and if nature includes animals, then rights-based claims could be made on behalf of animals using existing rights of nature doctrine and strategy.<@\color{blue}<highlight\_end>@><@\color{orange}<doc-sep>@>Every now and then, a concept emerges with the power to alter orthodoxies.\n<@\color{blue}<highlight\_start>@>In the legal world, one such concept is the "rights of nature," an idea that can-and should-encompass new protections for wild animals.<@\color{blue}<highlight\_end>@><@\color{orange}<doc-sep>@>How would such a legal system work, and could giving rights to nature help in the legal battle against climate change? A few case studies offer some insight. <@\color{blue}<highlight\_start>@>According to the "Rights of Nature" doctrine, an ecosystem is entitled to legal personhood status and as such, has the right to defend itself in a court of law against harms, including environmental degradation caused by a specific development project or even by climate change.<@\color{blue}<highlight\_end>@>

<@\color{red}Output:@> 
In contrast to environmental rights, it is only a fairly recent phenomenon that assigns "constitutional significance to the experiences of individual nonhuman animals."<@\color{blue}<highlight\_separator>@>At the constitutional level, Ecuador was the first to recognize the rights of nature.<@\color{orange}<doc-sep>@>Looking at legislation, the recognition of animal sentience is another emerging trend, although this does not go so far as to establish rights for animals.<@\color{blue}<highlight\_separator>@>To assert these rights, giving rivers legal personhood helps protect them as they can then be represented in court.<@\color{blue}<highlight\_separator>@>...<@\color{instructionsColor}<cluster\_separator>@>At the most fundamental level, if nature has rights, and if nature includes animals, then rights-based claims could be made on behalf of animals using existing rights of nature doctrine and strategy.<@\color{instructionsColor}<cluster\_separator>@>In the legal world, one such concept is the "rights of nature," an idea that can-and should-encompass new protections for wild animals.<@\color{instructionsColor}<cluster\_separator>@>According to the "Rights of Nature" doctrine, an ecosystem is entitled to legal personhood status and as such, has the right to defend itself in a court of law against harms, including environmental degradation caused by a specific development project or even by climate change.

\end{mylisting}
% \newtcbinputlisting[caption=Example prompt provided to GPT-4., frame=tlrb, captionpos=b]{figures/prompt.txt}
\caption{Example input and output for the fine-tuned sentence planning model in the LFQA setting. Highlights separators are colored blue and cluster separators are colored green.}
\label{fig:ft_lfqa_planning}
\end{figure*}

\begin{figure*}[t]

\lstdefinestyle{promptStyle}
{
    basicstyle={\footnotesize\ttfamily},% footnotesize acceptable for monospace
    numbers=left,numberstyle=\footnotesize,
    xleftmargin=2.8em,% show line numbers, remove this entire line if you don't want the numbers.
    xrightmargin=1.5em,
    showstringspaces=false,
      showspaces=false,
        showtabs=false,
    tabsize=2,
    breaklines=true,
        flexiblecolumns=true,
        escapeinside={<@}{@>},
          breakatwhitespace=true
}

\newtcblisting{mylisting}[1]{
  enhanced,
  listing only,
  boxrule=0.8pt,
  sharp corners=downhill,
  top=0mm,
  bottom=0mm,
  left=2mm,
  right=0mm,
  boxsep=0mm,
  colframe=black,
  colback=white,
  listing options={
    style=#1
  }
}

\definecolor{instructionsColor}{rgb}{0.1, 0.5, 0.1}

\begin{mylisting}{promptStyle}
<@\color{red}Input:@> 
Q: 'Wild law': if animals have rights, can nature have rights too? 
<@\color{orange}<query\_documents\_separator>@>
Most constitutions address the environment, and the typical phrasing is anthropocentric: a human right to a healthy environment as seen, for example, in article 42 of the Constitution of Kenya: "Every person has the right to a clean and healthy environment . . . ." Newer trends adopt ecocentric or biocentric approaches and grant rights to nature (or its component parts, such as a river) at the constitutional or legislative level or through judicial decisions. <@\color{blue}<highlight\_start>@>In contrast to environmental rights, it is only a fairly recent phenomenon that assigns "constitutional significance to the experiences of individual nonhuman animals."<@\color{blue}<highlight\_end>@> Ecocentric or biocentric approaches that lodge a right in nature or its component parts also may be promising for the development of legally recognized animal rights. Rights of nature are not widespread, but they have potential for growth and impact. <@\color{blue}<highlight\_start>@>At the constitutional level, Ecuador was the first to recognize the rights of nature.<@\color{blue}<highlight\_end>@><@\color{orange}<doc-sep>@>Mainly, they have helped in cases of mistreatment of chimpanzees and elephants in captivity through petitioning for writs of habeas corpus in challenging the detainment of these animals. One of their longest running cases concerns Happy , an elephant kept alone in the Bronx Zoo (since 1977) who was ruled in June to not be a legal person, though two dissenters stressed Happy's rights of liberty and autonomy. <@\color{blue}<highlight\_start>@>Looking at legislation, the recognition of animal sentience is another emerging trend, although this does not go so far as to establish rights for animals.<@\color{blue}<highlight\_end>@> Put into action, a focus is on giving ecosystems legal guardianship and personhood. One example of this is with rights of rivers. <@\color{blue}<highlight\_start>@>To assert these rights, giving rivers legal personhood helps protect them as they can then be represented in court.<@\color{blue}<highlight\_end>@> <@\color{blue}<highlight\_start>@>This has occurred with all the rivers in Bangladesh and the Atrato River in Columbia, and many others.<@\color{blue}<highlight\_end>@><@\color{orange}<doc-sep>@> "I had to choose a single sort of leader in the rights of nature movement; it would definitely be indigenous governments in the United States," Bradshaw says. But it took many years for these ideas to gain official consideration outside tribal courts and Western judicial systems. <@\color{blue}<highlight\_start>@>In 1972, Supreme Court Justice William O. Douglas first asserted the legal rights of nature in a U.S. court in a dissenting opinion on the Sierra Club v. Morton case.<@\color{blue}<highlight\_end>@> Colombian officials planned to kill the hippos, but an attorney filed a lawsuit on behalf of them in 2020, hoping to save them from euthanasia. Experts in the U.S. applied to provide testimony in the Colombian court. <@\color{blue}<highlight\_start>@>Earlier this year, the U.S. District Court for the Southern District of Ohio granted the application , thereby officially recognizing animals - hippos - have legal rights akin to humans for the first time in U.S. history.<@\color{blue}<highlight\_end>@>

<@\color{red}Output:@> 
The idea of extending legal rights to non-human entities, including animals and nature, is not new.

\end{mylisting}
% \newtcbinputlisting[caption=Example prompt provided to GPT-4., frame=tlrb, captionpos=b]{figures/prompt.txt}
\caption{Example input and output for the fine-tuned sentence-by-sentence generation in the LFQA setting. This is the first sentence generated in the summary, so there is no prefix.}
\label{fig:ft_lfqa_sentence_generation}
\end{figure*}

%% file: tables/tab_autoais_sent.tex
\begin{table}[t!]
    \centering
    \small
    \adjustbox{max width=\columnwidth}{
        \begin{tabular}{llcc}
            & Method & \multicolumn{1}{c}{\textsc{AutoAIS$\uparrow$}} & \multicolumn{1}{c}{\textsc{AutoAIS$_{full\ sentence}$$\uparrow$}} \\
             \toprule
              \parbox[t]{2mm}{\multirow{3}{*}{\rotatebox[origin=c]{90}{ICL}}} & \textsc{ALCE} & 88.7 & 88.7 \\
             & \textsc{Attr. First} & 79.5 & 80.6  \\
             & \textsc{Attr. First}\textsubscript{\texttt{CoT}} & 72.8 & 75.0 \\
             \midrule
             \parbox[t]{2mm}{\multirow{2}{*}{\rotatebox[origin=c]{90}{FT}}} & \textsc{Attr. First} & 64.4 & 80.1 \\
             & \textsc{Attr. First}\textsubscript{\texttt{joint}} & 48.9 & 83.9 \\

             \bottomrule
        \end{tabular}
    }
    % \vspace{-4pt}
    \caption{MDS results comparing our reported AutoAIS to a variant of AutoAIS where the full-sentences containing the highlights are used as premise. The difference between the two AutoAIS variants is that the former might contain non-grammatical sentences.}
    \label{tab:autoais_sent_mds}
\end{table}

\begin{table}[t!]
    \centering
    \small
    \adjustbox{max width=\columnwidth}{
        \begin{tabular}{llcc}
             & Method & \multicolumn{1}{c}{\textsc{AutoAIS$\uparrow$}} & \multicolumn{1}{c}{\textsc{AutoAIS$_{full\ sentence}$$\uparrow$}} \\
             \toprule
             \parbox[t]{2mm}{\multirow{3}{*}{\rotatebox[origin=c]{90}{ICL}}} & \textsc{ALCE} & 49.8 & 49.8 \\
             & \textsc{Attr. First} & 78.7 & 80.9 \\
             & \textsc{Attr. First}\textsubscript{\texttt{CoT}} & 89.3 & 91.5 \\
             \midrule
            \parbox[t]{2mm}{\multirow{2}{*}{\rotatebox[origin=c]{90}{FT}}} & \textsc{Attr. First} & 52.7 & 56.0\\
             & \textsc{Attr. First}\textsubscript{\texttt{joint}} & 44.9 & 69.2 \\
             \bottomrule
        \end{tabular}
    }
    % \vspace{-4pt}
    % \caption{Results for the long-form QA dataset derived from \citet{liu-etal-2023-evaluating} annotations.}
    \caption{LFQA results comparing our reported AutoAIS to a variant of AutoAIS where the full-sentences containing the highlights are used as premise. The difference between the two AutoAIS variants is that the former might contain non-grammatical sentences.}
    \vspace{-0.3cm}
    \label{tab:autoais_sent_qa}
\end{table}

%% file: tables/tab_example_predictions.tex
\begin{table*}
    \centering
    \small
    \adjustbox{max width=\textwidth}{
        \begin{tabular}{p{0.3cm}|p{14.7cm}}
            & Prediction \\
             \toprule
              \parbox[t]{2mm}{\multirow{2}{*}{1}} & \textbf{Generated sentence:} Hawaii's tourism industry is worth \$14 billion, and hotels and resorts are offering special packages to attract same-sex couples. [1]\\
              \\
              & \textbf{[1]} \ldots There's a large number of firms that are specialized in the marriage business."
\hl{In Hawaii, tourism is a \$14 billion industry}. Now, the island chain is positioning itself for a spike in visitors - among them, Honolulu-based hotel chain Aqua Hospitality. \hl{It already offers LGBT travel deals, including the "out and proud" package and the "civil unions romance special."}
Bigger beachfront resorts are also getting in on the action. The Sheraton Waikiki is the only hotel on Oahu providing on-site marriage licenses. General Manager Kelly Sanders says now the hotel is running a new ad campaign in mainland LGBT publications \ldots \\

        \midrule
              \parbox[t]{2mm}{\multirow{2}{*}{2}} & \textbf{Generated sentence:} Patten provided lobbying and consulting services to a Ukrainian oligarch and his political party, but failed to register as a foreign agent as required by law. [1,2]\\
              \\
              & \textbf{[1]} \ldots From 2014, \hl{Patten provided a "prominent" Ukrainian oligarch who isn't named in court papers and his Opposition Bloc political party with lobbying and consulting services}, according to the criminal information. A company Patten co-owned with a Russian national received more than \$1 million for the work, the U.S. said. Details about the Ukrainian oligarch match those of Sergei Lyovochkin, an Opposition Bloc leader whom prosecutors have previously identified as funding Manafort's work in Ukraine. He couldn't immediately been reached for comment. \hl{As part of his lobbying work, Patten violated the Foreign Agents Registration Act by not disclosing the work to the U.S.,} prosecutors said. No date has been set for his sentencing \ldots \\
 \\
              & \textbf{[2]} \ldots Prosecutors say Patten, who formed a consulting company with a person identified only as "Foreigner A," worked to set up meetings with members of Congress and also drafted talking points for Capitol Hill meetings.
\hl{The goal, prosecutors say, was to influence U.S. policy, but they say Patten never filed under the Foreign Agents Registration Act.} The law is aimed at promoting transparency about lobbying efforts in the United States.
 \ldots
 \\

        \midrule
              \parbox[t]{2mm}{\multirow{2}{*}{3}} & \textbf{Generated sentence:} On Tuesday, a shooting at a Home Depot store in North Dallas left three people injured: two police officers, Crystal Almeida and Rogelio Santander, and a Home Depot loss-prevention officer, Scott Painter. [1,2,3]\\
              \\
              & \textbf{[1]} \ldots \hl{The wounded officer is Crystal Almeida, 26. The third victim was identified as Scott Painter, a Home Depot loss-prevention officer. Almeida and Painter were still in critical condition on Wednesday}, but they were making "remarkable recoveries," Dallas Police Chief U. Renee Hall said at a news conference Wednesday morning. \ldots \\
              \\

              & \textbf{[2]} \ldots Along with Officer Santander, \hl{Officer Crystal Almeida and the Home Depot security guard, now identified as Scott Painter, all underwent surgery after the shooting.} Officer Almeida and Mr. Painter both remain hospitalized.
During the press conference Chief Hall said, "We are happy to report that Officer Crystal Almeida and our loss-prevention officer Scott Painter is making remarkable recovery. They are still in critical condition but we are optimistic about what we're seeing with them right now."
\hl{The man who police say shot the officers and the security guard - Armando Luis Juarez - woke up behind bars today.} After the shooting, a manhunt, and late night police chase, Juarez was taken into custody and ultimately transferred to the Dallas County Jail very early Wednesday morning.
\hl{Officers Santander and Almeida both joined the Dallas Police Department there years ago, assigned to the Northeast Division.}
 \ldots \\

              & \textbf{[3]} \ldots Mayor Mike Rawlings was presiding over a city council meeting when he announced the death of Rogelio Santander, a member of the police force for three years. \hl{Santander, officer Crystal Almeida and a loss-prevention officer for Home Depot were shot Tuesday by a man identified by police as 29-year-old Armando Luis Juarez. The two officers and the store loss-prevention officer underwent surgery for their injuries after the shooting in the north of the city}, Dallas Police Chief U. Renee Hall said late Tuesday.
\hl{Almeida and the loss-prevention officer, who hasn't been identified, were in critical condition Wednesday.}
 \ldots \\

             \bottomrule
        \end{tabular}
    }
    % \vspace{-4pt}
    \caption{Attribution examples for several sentences generated by the \textsc{Attr. First}\textsubscript{\texttt{CoT}} variant for the MDS settings.}
    \label{tab:example_predictions}
\end{table*}

%% file: tables/tab_partially_attributed_examples.tex
\begin{table*}
    \centering
    \small
    \adjustbox{max width=\textwidth}{
        \begin{tabular}{p{0.3cm}|p{14.7cm}}
            & Prediction \\
             \toprule
              \parbox[t]{2mm}{\multirow{2}{*}{1}} & \textbf{Output:} Despite economic challenges and \textbf{low approval ratings}, Obama's campaign finance report shows he can still attract major cash.\\
              \\
              & \textbf{Attribution:} \ldots \hl{The president's campaign finance report shows he can still pull in major cash despite a stagnant economy}, dipping approval ratings and grumblings among some liberal supporters that he has not done enough for their cause. \ldots \\

        \midrule
              \parbox[t]{2mm}{\multirow{2}{*}{2}} & \textbf{Output:} \textbf{The Sheraton Waikiki is the only hotel on Oahu that provides on-site marriage licenses}, and it is running a new ad campaign in mainland LGBT publications.\\
              \\
              & \textbf{Attribution:} \ldots  The Sheraton Waikiki is the only hotel on Oahu providing on-site marriage licenses. \hl{General Manager Kelly Sanders says now the hotel is running a new ad campaign in mainland LGBT publications.} \ldots \\

        \midrule
              \parbox[t]{2mm}{\multirow{2}{*}{3}} & \textbf{Output:} Tarek and his wife Christina tried for two years to have a son, going through two failed IVF attempts and \textbf{a miscarriage} before welcoming their son Brayden James in August.\\
              \\
              & \textbf{Attribution:} \ldots \hl{The couple, who are already parents to daughter Taylor, tried for more than two years to get pregnant with their son, going through two failed attempts at IVF} - including one resulting in a miscarriage at eight weeks, just before Tarek's diagnosis. \hl{Now in remission, the new dad, who welcomed son Brayden James with Christina, 32, in August, is positive about his family's future."} \ldots \\

             \bottomrule
        \end{tabular}
    }
    % \vspace{-4pt}
    \caption{Example predictions of the \textsc{Attr. First}\textsubscript{\texttt{CoT}} which were annotated with partial support. The information inferred from the words in bold is not supported by the highlights provided by the model, but is supported by sentences in close proximity. }
    \label{tab:partially_attributed_examples}
\end{table*}

%% file: tables/tab_prefix_examples.tex
\begin{table*}
    \centering
    \small
    \adjustbox{max width=\textwidth}{
        \begin{tabular}{p{0.3cm}|p{14.7cm}}
            & Prediction \\
             \toprule
              \parbox[t]{2mm}{\multirow{2}{*}{1}} & \textbf{w/ prefix:} The Ohio River starts at the confluence of the Allegheny and Monongahela Rivers in Pittsburgh, Pennsylvania, and flows 981 miles to its ending point in Cairo, Illinois. \textbf{The river} flows in a southwesterly direction from western Pennsylvania to its mouth on the Mississippi River at the southern tip of Illinois.\\
              \\
              & \textbf{w/o prefix:} The Ohio River starts at the confluence of the Allegheny and Monongahela Rivers in Pittsburgh, Pennsylvania, and flows 981 miles to its ending point in Cairo, Illinois. \textbf{The Ohio River} is a 981-mile (1,579 km) long river in the United States. \\

        \midrule
            \parbox[t]{2mm}{\multirow{2}{*}{2}} & \textbf{w/ prefix:} Idina Menzel and Taye Diggs, who starred together in the Broadway musical "Rent," have decided to end their marriage. \textbf{Their} main concern is their son. \textbf{The couple} , who met while starring in the Broadway musical "Rent," have a 4-year-old son, Walker Nathaniel. \textbf{Diggs and Menzel} , both 42, met as costars in the original production of the Broadway musical Rent in the mid-1990s. They worked at their marriage, going through tough times like everyone else, but they loved each other very much. \\
            \\
            & \textbf{w/o prefix:} Idina Menzel and Taye Diggs, who have been married for 10 years, have a 4-year-old son named Walker. Idina Menzel and Taye Diggs  have asked for privacy during this time, as their main concern is for their son. \textbf{Taye Diggs and Idina Menzel}  have decided to separate after ten years of marriage, but their main concern is for their son. \textbf{Taye Diggs and Idina Menzel} , who met in the 1990s while starring in the Broadway musical Rent, have asked for privacy during this difficult time. \textbf{Taye Diggs and Idina Menzel} , who met in the mid-1990s while starring in the original production of the Broadway musical Rent, married in 2003 and had a son, Walker, in September 2009. \\

        \midrule
            \parbox[t]{2mm}{\multirow{2}{*}{3}} & \textbf{w/ prefix:} By exposing students to different religious texts, practices, cultures, and social teachings, they can gain a deeper understanding of the world around them and develop critical thinking skills. \textbf{However ,} some Americans believe that teaching religious doctrine in public schools infringes on the First Amendment right to free exercise of religion. \\
            \\
            & \textbf{w/o prefix:} By exposing students to different religious texts, practices, cultures, and social teachings, they can gain a deeper understanding of the world around them and develop critical thinking skills. Some Americans believe that excluding religious sentiment from public schools infringes on the First Amendment right to free exercise of religion. \\

             \bottomrule
        \end{tabular}
    }
    % \vspace{-4pt}
    \caption{Example predictions of the ICL \textsc{Attr. First} model when prompted with prefix and without prefix. }
    \label{tab:prefix_examples}
\end{table*}

%% file: tables/token_count.tex
% \usepackage{graphicx}
\begin{table*}[t!]
\centering
\resizebox{\textwidth}{!}{%
\begin{tabular}{llcccccccc|cccc}
\hline
    &      & \multicolumn{2}{c}{Content Selection}                                                                              & \multicolumn{2}{c}{Generation Planning}                                                                         & \multicolumn{2}{c}{Sentence Fusion}                                                                                & \multicolumn{2}{c|}{CoT}                                                                                           & \multicolumn{2}{c}{End-to-end}                                                                                    & \multicolumn{2}{c}{ALCE}                                                                                         \\
    &      & Input                                                     & Output                                                 & Input                                                    & Output                                               & Input                                                      & Output                                                & Input                                                     & Output                                                 & Input                                                     & Output                                                & Input                                                    & Output                                                \\ \hline
\toprule
             \parbox[t]{2mm}{\multirow{2.5}{*}{\rotatebox[origin=c]{90}{\large \textbf{ICL}}}} & MDS  & \begin{tabular}[c]{@{}c@{}}6879\\ ($\pm$ 1893)\end{tabular}  & \begin{tabular}[c]{@{}c@{}}551\\ ($\pm$ 467)\end{tabular} & \begin{tabular}[c]{@{}c@{}}6425\\ ($\pm$ 1728)\end{tabular} & \begin{tabular}[c]{@{}c@{}}11\\ ($\pm$ 22)\end{tabular} & \begin{tabular}[c]{@{}c@{}}17003\\ ($\pm$ 13203)\end{tabular} & \begin{tabular}[c]{@{}c@{}}133\\ ($\pm$ 80)\end{tabular} & \begin{tabular}[c]{@{}c@{}}10273\\ ($\pm$ 1558)\end{tabular} & \begin{tabular}[c]{@{}c@{}}478\\ ($\pm$ 220)\end{tabular} & \begin{tabular}[c]{@{}c@{}}6879\\ ($\pm$ 1893)\end{tabular}  & \begin{tabular}[c]{@{}c@{}}177\\ ($\pm$ 55)\end{tabular} & \begin{tabular}[c]{@{}c@{}}7341\\ ($\pm$ 1974)\end{tabular} & \begin{tabular}[c]{@{}c@{}}193\\ ($\pm$ 44)\end{tabular} \\
    & LFQA & \begin{tabular}[c]{@{}c@{}}10727\\ ($\pm$ 3206)\end{tabular} & \begin{tabular}[c]{@{}c@{}}142\\ ($\pm$ 161)\end{tabular} & \begin{tabular}[c]{@{}c@{}}2536\\ ($\pm$ 1651)\end{tabular} & \begin{tabular}[c]{@{}c@{}}9\\ ($\pm$ 13)\end{tabular}  & \begin{tabular}[c]{@{}c@{}}3734\\ ($\pm$ 1784)\end{tabular}   & \begin{tabular}[c]{@{}c@{}}45\\ ($\pm$ 28)\end{tabular}  & \begin{tabular}[c]{@{}c@{}}2962\\ ($\pm$ 635)\end{tabular}   & \begin{tabular}[c]{@{}c@{}}127\\ ($\pm$ 89)\end{tabular}  & \begin{tabular}[c]{@{}c@{}}10727\\ ($\pm$ 3206)\end{tabular} & \begin{tabular}[c]{@{}c@{}}91\\ ($\pm$ 85)\end{tabular}  & \begin{tabular}[c]{@{}c@{}}8476\\ ($\pm$ 3809)\end{tabular} & \begin{tabular}[c]{@{}c@{}}95\\ ($\pm$ 89)\end{tabular}  \\ \hline
\toprule
             \parbox[t]{2mm}{\multirow{2.5}{*}{\rotatebox[origin=c]{90}{\large \textbf{FT}}}}  & MDS  & \begin{tabular}[c]{@{}c@{}}1661\\ ($\pm$ 116)\end{tabular}   & \begin{tabular}[c]{@{}c@{}}456\\ ($\pm$ 30)\end{tabular}  & \begin{tabular}[c]{@{}c@{}}1041\\ ($\pm$ 75)\end{tabular}   & \begin{tabular}[c]{@{}c@{}}456\\ ($\pm$30)\end{tabular} & \begin{tabular}[c]{@{}c@{}}348\\ ($\pm$ 7)\end{tabular}       & \begin{tabular}[c]{@{}c@{}}29\\ ($\pm$ 0.6)\end{tabular} & -                                                         & -                                                      & \begin{tabular}[c]{@{}c@{}}1661\\ ($\pm$ 116)\end{tabular}   & \begin{tabular}[c]{@{}c@{}}283\\ ($\pm$ 9)\end{tabular}  & -                                                        & -                                                     \\
    & LFQA & \begin{tabular}[c]{@{}c@{}}2883\\ ($\pm$162)\end{tabular}    & \begin{tabular}[c]{@{}c@{}}121\\ ($\pm$11)\end{tabular}   & \begin{tabular}[c]{@{}c@{}}348\\ ($\pm$36)\end{tabular}     & \begin{tabular}[c]{@{}c@{}}121\\ ($\pm$11)\end{tabular} & \begin{tabular}[c]{@{}c@{}}209\\ ($\pm$18)\end{tabular}       & \begin{tabular}[c]{@{}c@{}}27\\ (+ -1)\end{tabular}   & -                                                         & -                                                      & \begin{tabular}[c]{@{}c@{}}2883\\ ($\pm$ 162)\end{tabular}   & \begin{tabular}[c]{@{}c@{}}60\\ ($\pm$ 5)\end{tabular}   & -                                                        & -                                                     \\ \hline
\end{tabular}%
}
\caption{Average number of tokens  (and standard deviation) for each of the steps in our pipeline, as well as our two baselines.}
\label{tab:token_count}
\end{table*}

%% file: tables/tab_latency.tex
% \usepackage{graphicx}
\begin{table*}[h!]
\centering
\resizebox{0.8\textwidth}{!}{%
\begin{tabular}{llcccc|cc}
\hline
    &      & Content Selection & Generation Planning & Sentence Fusion & CoT        & End-to-end & ALCE \\ \hline
\toprule
             \parbox[t]{2mm}{\multirow{1.7}{*}{\rotatebox[origin=c]{90}{\textbf{ICL}}}} & MDS  & 33.1              & 5.2                 & 5.6             & 17.1       & 7.0        & 9.2  \\
    & LFQA & 6.3               & 4.7                 & 6.8             & 3.3        & 6.1        & 6.7  \\ \hline
\toprule
             \parbox[t]{2mm}{\multirow{1.7}{*}{\rotatebox[origin=c]{90}{\textbf{FT}}}}  & MDS  & 68.8              & 5.7                 & 2.4             & \textbf{-} & 1.7        & -    \\
    & LFQA & 10.6              & 3.4                 & 1.6             & \textbf{-} & 1.2        & -    \\ \hline
\end{tabular}%
}
\caption{Average time (in seconds) each of the steps in our pipeline takes per instance, as well as for our two baselines.}
\label{tab:latency}
\end{table*}

%% file: tables/tab_rarr.tex
% \usepackage{graphicx}
\begin{table}[t!]
\centering
\resizebox{\columnwidth}{!}{%
\begin{tabular}{lcc|ccc}
\hline
Task & \textsc{Rouge$_L$} & \textsc{BertScore} & \textsc{AutoAIS} & \textsc{Length} & \textsc{No Att. (\%)} \\ \hline
MDS  & 19.0   & 86.1      & 67.8    & 137.7  & 0.7     \\
LFQA & 33.1   & 89.6      & 50.2    & 149.0  & 9.6     \\ \hline
\end{tabular}%
}
\caption{Results of the post-generation attribution approach, adapted from the RARR algorithm, on the Multi-document summarization (MDS) and the Long-form Question-answering (LFQA) tasks.}
\label{tab:rarr_results}
\end{table}